\documentclass[onecolumn]{article}         
\usepackage[top=3.6cm, bottom=3.2cm, left=2.5cm, right=2.5cm]{geometry}
\usepackage{tabularx}
\usepackage{graphicx}
\usepackage{multirow}
\usepackage{multicol}
\usepackage{floatrow}
\usepackage{subfigure}
\usepackage{makecell,rotating}
\usepackage{textcomp,booktabs}
\usepackage[usenames,dvipsnames]{color}
\usepackage{colortbl}
\definecolor{setgray}{gray}{.98}
\definecolor{setblue}{gray}{.58}
\newcolumntype{C}[1]{>{\centering\arraybackslash}p{#1}}
\newcolumntype{L}[1]{>{\raggedright\arraybackslash}p{#1}}
\usepackage{amssymb}
\usepackage{amsmath}
\usepackage{bm}
\usepackage{cite}
\usepackage[colorlinks=true,linkcolor=red,citecolor=blue]{hyperref}
\usepackage{url}

\begin{document}
\renewcommand{\thefootnote}{\fnsymbol{footnote}}
\title{Land-Cover Classification with High-Resolution Remote Sensing Images Using Transferable Deep Models\footnote{A website is available at \url{https://x-ytong.github.io/project/GID.html}}}

\author{Xin-Yi Tong$^1$, Gui-Song Xia$^{1, 2}$, Qikai Lu$^3$, Huanfeng Shen$^4$,\\ Shengyang Li$^5$, Shucheng You$^6$, Liangpei Zhang$^1$
\vspace{3mm}
\\
$^1${\em State Key Laboratory LIESMARS, Wuhan University, China.}\\
$^2${\em School of Computer Science, Wuhan University, China.}\\
$^3${\em Electronic Information School, Wuhan University, China.}\\
$^4${\em School of Resource and Environmental Sciences, Wuhan University, China.}\\
$^5${\em Key Laboratory of Space Utilization, Tech. $\&$ Eng. Center for Space Utilization,}\\
    {\em Chinese Academy of Sciences, China.}\\
$^6${\em Remote Sensing Department, China Land Survey and Planning Institute, China.}\\
}
\date{}
\maketitle

\begin{abstract}
In recent years, large amount of high spatial-resolution remote sensing (HRRS) images are available for land-cover mapping. However, due to the complex information brought by the increased spatial resolution and the data disturbances caused by different conditions of image acquisition, it is often difficult to find an efficient method for achieving accurate land-cover classification with high-resolution and heterogeneous remote sensing images. In this paper, we propose a scheme to apply deep model obtained from labeled land-cover dataset to classify unlabeled HRRS images. The main idea is to rely on deep neural networks for presenting the contextual information contained in different types of land-covers and propose a pseudo-labeling and sample selection scheme for improving the transferability of deep models. More precisely, a deep Convolutional Neural Networks (CNNs) is first pre-trained with a well-annotated land-cover dataset, referred to as the {\em source data}. Then, given a {\em target image} with no labels, the pre-trained CNN model is utilized to classify the image in a patch-wise manner. The patches with high confidence are assigned with pseudo-labels and employed as the queries to retrieve related samples from the source data. The pseudo-labels confirmed with the retrieved results are regarded as supervised information for fine-tuning the pre-trained deep model. To obtain a pixel-wise land-cover classification with the target image, we rely on the fine-tuned CNN and develop a hybrid classification by combining patch-wise classification and hierarchical segmentation. In addition, we create a large-scale land-cover dataset containing $150$ Gaofen-2 satellite images for CNN pre-training. Experiments on multi-source HRRS images, including Gaofen-2, Gaofen-1, Jilin-1, Ziyuan-3, Sentinel-2A, and Google Earth platform data, show encouraging results and demonstrate the applicability of the proposed scheme to land-cover classification with multi-source HRRS images.
\end{abstract}

\section{Introduction}
\label{sec:introduction}
Land-cover classification with remote sensing (RS) images plays an important role in many applications such as land resource management, urban planning, precision agriculture, and environmental protection \cite{application1,application2,application3,application4,application5,application6}. In recent years, high-resolution remote sensing (HRRS) images are increasingly available. Meanwhile, multi-source and multi-temporal RS images can be obtained over different geographical areas \cite{source}. Such large amount of heterogeneous HRRS images provide detailed information of the land surface, and therefore open new avenues for large-coverage and multi-temporal land-cover mapping.
However, the rich details of objects emerging in HRRS images, such as the geometrical shape and structural content of objects, bring more challenges to land-cover classification~\cite{separability}.
Furthermore, diverse imaging conditions usually lead to photographic distortions, variations in scale and changes of illumination in RS images, which often seriously reduces the separability among different classes~\cite{DAreview}.
Due to these influences, optimal classification models learned from certain annotated images always quickly lose their effectiveness on new images captured by different sensors or by the same sensor but from different geo-locations. Therefore, it is intractable to find an efficient and accurate land-cover classification method for HRRS images with large diversities.

To characterize the image content of different land-cover categories, many methods investigated the use of spectral and spectral-spatial features to interpret RS images \cite{spectral1,spectral2,spectral3,spatialSpectral1,spatialSpectral2,spatialSpectral3,spatialSpectral4,spatialSpectral5}. However, due to the detailed and structural information brought by the gradually increased spatial resolution, the spectral and spectral-spatial features have difficulty in describing the contextual information contained in the images \cite{semanticGap1,semanticGap2,hu2016fast,yu2016color}, which are often essential in depicting land-cover categories in HRRS images.
Recently, it has been reported that effective characterization of contextual information in HRRS images can largely improve the classification performance~\cite{shao2013extreme,hu2017deep,yang2015learning}.
Among them, deep Convolutional Neural Networks (CNNs) have been drawn much attention in the understanding of HRRS images~\cite{SceneClassification1,deepReview},
mainly because of their strong capability to depict high-level and semantic aspects of images~\cite{AlexNet,visualizing}. Currently, various deep models have been adopted to cope with challenging issues in RS image understanding, including {\em e.g.} scene classification~\cite{SceneClassification1,SceneClassification2}, object detection~\cite{ObjectDetection}, image retrieval~ \cite{Retrieval,Jiang2017,XiaTHZDZ17}, as well as land-cover classification~\cite{deepFeature1,deepFeature2,deepFeature3,endToEnd1,endToEnd2,endToEnd3}.

Nevertheless, there are two main problems in applying deep model to land-cover classification with multi-source HRRS images, which are listed below.

\begin{itemize}
\item[-]\emph{The inadequate transferability of deep learning models}: Due to the diverse distributions of objects and spectral shifts caused by the different acquisition conditions of images, deep models trained on a certain set of annotated RS images may not be effective when dealing with images acquired by different sensors or from different geo-locations~\cite{domainAdaption}. To obtain satisfactory land-cover classification on a RS image of interest, referred as the {\em target image}, new specific annotated samples closely related to it are often necessary for model fine-tuning~\cite{endToEnd1}. Nevertheless, considering that manual annotation requires high labor intensity and is often time-consuming, it is infeasible to label sufficient samples for continuously accumulated multi-source RS images~\cite{lu2017active,hu2015comparative}.
\item[-]\emph{The lack of well-annotated large-scale land-cover dataset}: The identification capability of CNN models relies heavily on the quality and quantity of the training data \cite{batchSelection}. Up to now, several land-cover datasets have been proposed in the community, and have advanced a lot deep-learning-based land-cover classification approaches~\cite{dataset1,dataset2,dataset3}. However, the geographic areas covered by most of existing land-cover datasets~\cite{DatasetReview,dataset1,dataset3} do not exceed $10 km^{2}$ and somewhat similar in geographic distributions~\cite{dataset4}. The lack of variations in geographic distributions of annotated HRRS images may cause overfitting in model training and limit the generalization ability of learned models. Overall, the insufficient or unqualified training data restrict the availability of deep models for HRRS images.
\end{itemize}

In this paper, we propose a scheme to adapt deep models to land-cover classification with multi-source HRRS images, which don't have any labeling information. Considering that the textures and structures of the objects are not affected by the spectral shifts, we use contextual information extracted by CNN to automatically mine samples for deep model fine-tuning. Concretely, unlabeled samples in the target image are identified by a CNN model pre-trained on an annotated HRRS dataset, which is referred to as the {\em source data}. A subset of them with high confidence are assigned with pseudo-labels and employed to retrieve similar samples from the source data. Finally, the returned results are used to determine whether the pseudo-labels are reliable. In our classification process, a patch-wise classification is initially conducted on the image relying on the multi-scale contextual information extracted by CNN. Then, a hierarchical segmentation is used for obtaining the object boundary information, which is integrated into the patch-wise classification map for accurate results. Specifically, for pre-training CNN models, we annotate $150$ Gaofen-2 satellite images to construct a land-cover classification dataset, which is named after {\em Gaofen Image Dataset} (GID).

In summary, the contributions of this paper are as follows:
\begin{itemize}
\vspace{-2mm}
\item[-]We propose a scheme to train transferable deep models, which enables one to achieve land-cover classification by using unlabeled multi-source RS images with high spatial resolution. In addition, we develop a hybrid land-cover classification that can simultaneously extract accurate category and boundary information of HRRS images. Experiments conducted on multi-source HRRS images, including Gaofen-2, Gaofen-1, Jilin-1, Ziyuan-3, Sentinel-2A, and Google Earth platform data obtain promising results and demonstrate the effectiveness of the proposed scheme.
\vspace{-2mm}
\item[-]We present a large-scale land-cover classification dataset, namely GID, which is consist of $150$ high-resolution Gaofen-2 images and covers areas more than $50,000$ $km^{2}$ in China.
To our knowledge, GID is the first and largest well-annotated land-cover classification dataset with high-resolution remote sensing images up to $4$ meters. It can provide the research community a high-quality dataset to advance land-cover classification with HRRS images, {\em like} Gaofen-2 imagery.
\end{itemize}

A preliminary version of this work was presented in~\cite{tong2018large}.

The remainder of the paper is organized as follows: In Section \ref{sec:related work}, we introduce the related works. In Section \ref{sec:methodology}, the introduction of our land-cover classification algorithm is presented. In Section \ref{sec:data}, the details and properties of GID coupled with other examined images are described. We present the results of experiments and sensitivity analysis in Section \ref{sec:experiments} and Section \ref{sec:sensitivity analysis}, and give the discussion in Section \ref{sec:discussion}. Finally, we conclude our work in Section \ref{sec:conclusion}.

\section{Related work}
\label{sec:related work}
\textbf{Land-cover Classification:} Land-cover classification with RS images aims to associating each pixel in a RS image with a pre-defined land-cover category. To this end, classification approaches using spectral information have been intensively studied. These methods can interpret RS images using the spectral features of individual pixels~\cite{spectral1,spectral2,spectral3}, but their performance is often heavily affected by intra-class spectral variations and noises~\cite{spectralDrawback1,spectralDrawback2,spectralDrawback3}. More recently, the spatial information has been taken into consideration for land-cover classification \cite{spatialSpectral1,spatialSpectral2,spatialSpectral3,spatialSpectral4,spatialSpectral5}. Spectral-spatial classification incorporates spatial information, such as texture \cite{texture,XiaDG10,XiaLBZ17}, shape~\cite{shape}, and structure features~\cite{structure,Xia2010StructuralHS}, to improve the separability of different categories in the feature space. It has been reported that spectral-spatial approaches can effectively boost the categorization accuracy compared with the methods using spectral features \cite{comparison1,comparison2,comparison3,comparison4}.
However, with the improvement of spatial-resolution of RS images, rather than discriminating in spectral or spectral-spatial information of local pixels, land-cover types are more categorized in contextual information and spatial relationship of ground objects~\cite{shao2013extreme,yang2015learning}.

Recently, deep neural network models have been intensively studied to address the task of land-cover classification and reported impressive performance, see {\em e.g.}~\cite{deepFeature1,deepFeature2,deepFeature3,deepFeature4,deepFeature5,deepFeature6}. In contrast with conventional methods that employ spectral or spectral-spatial features for land-cover description, a significant advantage of deep learning approaches is that they are able to adaptively learn discriminative features from images~\cite{visualizing}. Land-cover classification approaches that utilize deep features treat CNN models as feature extractors and employ conventional classifiers \cite{deepFeature1,deepFeature2,deepFeature4,deepFeature5,deepFeature6}, such as support vector machine (SVM) and logistic regression, for classification. As an alternative, end-to-end CNN models are adopted to interpret RS images \cite{endToEnd1,endToEnd2,endToEnd3,endToEnd4,endToEnd5}. End-to-end CNN models, such as Fully Convolutional Networks (FCN) \cite{endToEnd1}, conduct dense land-cover labeling for RS images without using additional classifiers or post-processing. Although, compared with utilizing deep features, end-to-end CNN models are more efficiency for classification, the receptive field in CNNs leads to the loss of fine resolution detail~\cite{endToEnd5}. To address this problem, an effective solution is to replace down-sampling process with the structure that preserves spatial information~\cite{WODS1,WODS2}, such as the max pooling indices applied in SegNet~\cite{Segnet}, and the dilated convolutions employed in DeepLab~\cite{DeepLab}. Moreover, the detailed spatial information can be obtained through complementary classification frameworks. The ensemble MLP-CNN classifier ~\cite{deepFeature3} fuses results acquired from the CNN based on deep spatial feature representation and from the multi-layer perceptron (MLP) based on spectral discrimination, which compensates the uncertainty in object boundary partition. And object-based convolutional neural network (OCNN) ~\cite{OCNN} incorporates CNN into the framework of object-based image analysis for more precise object boundaries and achieves encouraging performance in high-resolution urban scenes.

\textbf{Transfer Learning:} In practical land-cover classification applications, the available ground-truth samples are usually not sufficient in number and not adequate in quality for training a high-performance classifier. Thus, to improve the classification accuracy, transfer learning has been adopted as a promising solution \cite{DAreview}. Transfer learning aims to adapt models trained to solve a specific task to a new, yet related, task \cite{TLdefine}. The existing task is usually referred to as \emph{source domain} and the new task is \emph{target domain}. Two major categories of transfer learning approaches have been studied in the RS community: supervised learning methods and semi-supervised learning methods. Supervised learning approaches assume that the training set is available for both the source and target domain. They are commonly based on the selection of invariant features \cite{InvariantFeature1,InvariantFeature2}, the adaptation of data distributions \cite{DistributionAdaptation1,DistributionAdaptation2,DistributionAdaptation3}, and active learning \cite{ActiveLearning1,ActiveLearning2,ActiveLearning3}. By contrast, the approaches are defined as semi-supervised if they use only unlabeled samples of the target domain. Semi-supervised learning methods exploit the structural information of unlabeled samples in the feature space to better model the distribution of classes \cite{SSLdefine}. They are effective in a wide range of situations and do not require a strict match between the source and target domains \cite{ClassifierAdaptation1,ClassifierAdaptation2,ClassifierAdaptation3}. However, their performance is extremely dependent on the classifier's ability to learn structural information of the target domain.

On the other hand, deep neural networks are widely used for transfer learning in recent years due to their ability to model high-level abstractions of data \cite{DL}. CNN trained on large-scale natural image dataset have been transferred to interpret RS images, either by directly using the pre-trained network as a feature extractor \cite{SceneClassification1,DeepFE1,DeepFE2}, or fine-tuning the network with large-scale RS dataset \cite{SceneClassification2}. However, due to the huge number of model parameters, deep models require large amount of supervised information of the target domain to avoid the over-fitting problem \cite{JointFinetune2}. To reduce the required number of training samples, semi-transfer deep convolutional neural networks (STDCNN) is proposed for land-use mapping \cite{STDCNN}. STDCNN fuses a small CNN designed to analyze RS images and a deep CNN used for domain adaptation. It achieves promising performance with a small amount of labeled target samples.

\section{Methodology}
\label{sec:methodology}
To efficiently conduct land-cover classification with multi-source HRRS images, we propose a scheme to train transferable deep models, which is pre-trained on labeled land-cover dataset and can be applied to unlabeled HRRS images. Assume that there is a well-annotated large-scale dataset and a newly acquired image without labeling information. We define two domains, called \emph{source domain} $\textbf{\emph{D}}_{\textbf{\emph{S}}}$ and \emph{target domain} $\textbf{\emph{D}}_{\textbf{\emph{T}}}$ that are separately associated with the labeled and unlabeled images. Our aim is to exploit the information learned from the source domain $\textbf{\emph{D}}_{\textbf{\emph{S}}}$ to conduct classification in the target domain $\textbf{\emph{D}}_{\textbf{\emph{T}}}$.

Firstly, we use $\textbf{\emph{D}}_{\textbf{\emph{S}}}$ to pre-train a deep model specific to RS domain, which is presented in Section \ref{sec:training}. Given a target image $\chi_{\textbf{\emph{T}}}$ belonging to $\textbf{\emph{D}}_{\textbf{\emph{T}}}$, we divide it into patches $\textbf{U}={\{\textbf{x}_{i}\}}^{I}_{i=1}$ with non-overlapping grid partition. Our method automatically searches reliable training samples from $\textbf{U}$ to learn transferable deep model for $\chi_{\textbf{\emph{T}}}$, as introduced in Section \ref{sec:fine-tuning}. Subsequently, we utilize the fine-tuned deep model to classify $\textbf{x}_{i}$ for all $i\in\{1,\ldots,I\}$. Our classification scenario combines patch-wise categorization and object-based voting, which is described in Section \ref{sec:classification}.

\subsection{Learning deep model for land-cover classification}
\label{sec:training}
CNN models are deep hierarchical architectures which commonly consist of three main types of layers: convolutional layers, pooling layers, and fully-connected layers. Convolutional layers perform as hierarchical feature extractors, pooling layers conduct spatial down-sampling of feature maps, while fully-connected layer serve as the classifier to generate the predictive classification probabilities of the input data. In addition to these main layers, Residual Networks (ResNet) \cite{ResNet} adopt residual connections to improve the model performance. The structure of residual connection can greatly reduce the optimization difficulty, as well as enable the training of much deeper networks.

ResNet models have 5 versions, separately with 18, 34, 50, 101, and 152 layers. Compared to the models with shallow architecture, i.e. ResNet-18 and ResNet-34, ResNet-50 can achieve better classification performance. Compared to the models with very deep architecture, i.e. ResNet-101 and ResNet-152, ResNet-50 has fewer parameters and higher computational efficiency. Therefore, for a trad-off consideration between simplicity and computational efficiency, we employ ResNet-50 as the classifier in our work.

\begin{figure*}[htb!]
\centering
\includegraphics[width=1\textwidth]
{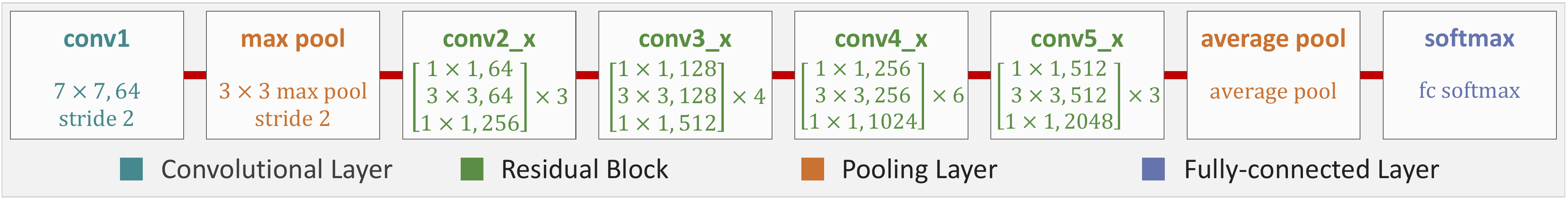}
\caption{The structure of ResNet-50. Different structures are represented by different colors.}
\label{figure:resnet}
\end{figure*}

ResNet-50 consists of 16 residual blocks, each of which has 3 convolutional layers that constitute a shortcut connection. The first convolutional layer of the overall model is followed by a max pooling layer. An average pooling layer, a full-connected layer, and a softmax layer are subsequent to the last convolutional layer. Fig. \ref{figure:resnet} shows the detailed structure of ResNet-50 we employed. Note that the default input size of ResNet-50 is $224\times 224\times 3$. To transfer deep model to classify images with only R, G, B bands (Jilin-1 satellite images, Google Earth platform data), we utilize ResNet-50 with 3 input channels. To classify images have R, G, B, NIR bands (Gaofen-1, Sentinel-2A, Ziyuan-3 satellite images), we adjust the input size of the conversional ResNet-50 to 4 channels. The difference between 3-channel and 4-channel models is that the kernel sizes in conv1 are $7\times 7\times 4$ and $7\times 7\times 3$, respectively. The rest of structures in the models are the same.

To pre-train a CNN model with strong discrimination ability for HRRS images, we construct a large-scale land-cover dataset, Gaofen Image Dataset (GID), which contains 150 well-annotated Gaofen-2 images and will be represented in Section \ref{sec:data}. The training images of GID are referred to as $\textbf{\emph{D}}_{\textbf{\emph{S}}}$ and are used to pre-train the ResNet-50 models.

\subsection{Learning transferable model for multi-source images}
\label{sec:fine-tuning}
Although CNNs have a certain degree of generalization ability, they are unable to achieve satisfactory classification results on multi-source RS images because of dramatic changes in acquisition conditions. To transfer CNN models for classifying RS images acquired under different conditions, we extract available information from the unlabeled target data to find a more accurate classification rule than using only labeled source data. Inspired by pseudo-label assignment \cite{PseudoLabel1,PseudoLabel2} and joint fine-tuning \cite{JointFinetune1,JointFinetune2} methods, we propose a semi-supervised transfer learning scheme to classify multi-source RS images. The main idea of pseudo-label assignment is to select valuable samples from the target domain based on the predicted classification confidence \cite{PseudoLabel1,PseudoLabel2}, however, the pseudo-labels may be unreliable. Joint fine-tuning optimizes the classification model by adding promising target samples into the source domain training set \cite{JointFinetune1,JointFinetune2}, however, it requires a small amount of labeled target samples. Our scheme combines the advantages of the aforementioned methods to acquire reliable training information from the unlabeled target domain for model optimization.

\begin{figure*}[htb!]
\centering
\includegraphics[width=0.8\textwidth]
{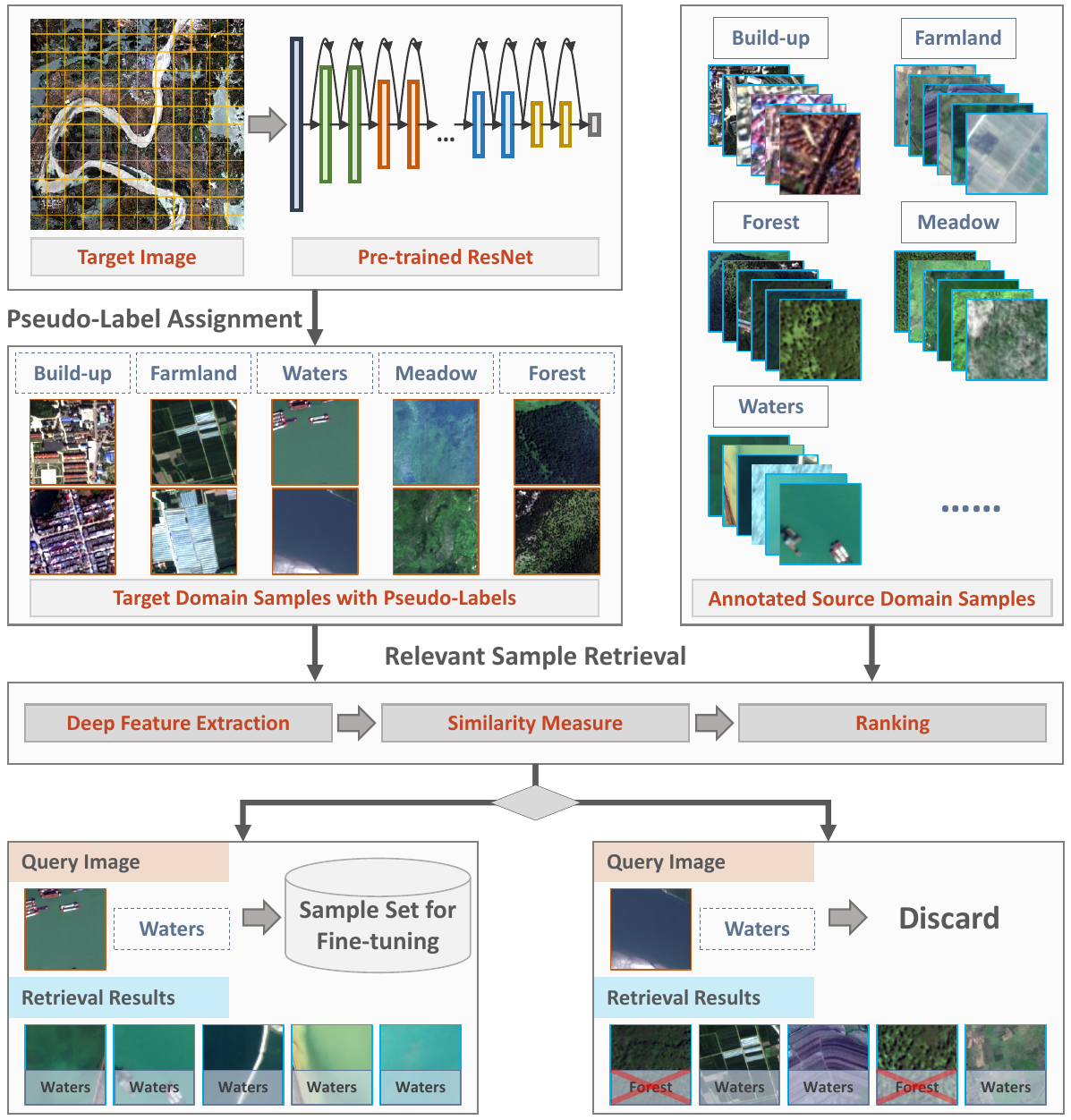}
\caption{Sample selection for model fine-tuning.}
\label{figure:selection}
\end{figure*}

As shown in Fig. \ref{figure:selection}, the proposed scheme is divided into two stages: pseudo-label assignment and relevant sample retrieval, which are presented in Section \ref{sec:assignment} and Section \ref{sec:retrieval}, respectively. Category predictions and deep features generated by CNN are used to search target samples that possess similar characteristics to the source domain. These relevant samples and their corresponding category predictions, which are referred to as pseudo-labels, are used for CNN model fine-tuning.

\subsubsection{Pseudo-label assignment}
\label{sec:assignment}
Given the patch set $\textbf{U}={\{\textbf{x}_{i}\}}^{I}_{i=1}$ of a target image $\chi_{\textbf{\emph{T}}}$, we input each patch $\textbf{x}_{i}$ to ResNet-50 that has been pre-trained on $\textbf{\emph{D}}_{\textbf{\emph{S}}}$. The output vectors of softmax layer form a set $\textbf{F}={\{\textbf{p}_{i}\}}^{I}_{i=1}$, where:

\begin{equation}
\textbf{p}_{i} = \{{p}_{i,1},{p}_{i,2},\ldots,{p}_{i,K}\}, \textbf{p}_{i}\in{\mathbb{R}}^{K}
\end{equation}
${p}_{i,k}$ represents the probability that patch $\textbf{x}_{i}$ belongs to class $k$, $k\in\{1,\ldots,K\}$, and $K$ is the total number of classes.

$\textbf{p}_{i}$ is the predicted classification probability vector, of which the highest probability value is $h=\mathop{\max}_{k\in\{1,\ldots,K\}}{p}_{i,k}$. Since CNN model has strong discriminating ability, we use the probability value to determine whether a sample is associated with a label. If the value of $h$ is greater than or equal to a threshold $\sigma$, the patch $\textbf{x}_{i}$ is reserved and assigned with a predicted category ${l}_{i}$. Otherwise, $\textbf{x}_{i}$ is removed from $\textbf{U}$. ${l}_{i}$ corresponds to the category represented by $h$ and is referred to as a pseudo-label. After removing all patches with low classification confidence from $\textbf{U}$, the remaining samples form a new set ${\textbf{U}}_{1}$.

\subsubsection{Relevant sample retrieval}
\label{sec:retrieval}
Considering that the pseudo-labels may be inaccurate, we search source samples that are similar to the selected target samples, and use the true-labels of the retrieved source samples to determine the reliability of the pseudo-labels. Assume that there are $J$ candidates remaining in the set ${\textbf{U}}_{1}$, ${\textbf{U}}_{1}={\{\textbf{x}_{j}\}}^{J}_{j=1}$. For the patch $\textbf{x}_{j}$, the information entropy $E_{j}$ is calculated to measure its classification uncertainty:

\begin{equation}
E_{j} = -\sum_{k=1}^{K}\\p_{j,k}\cdot log(p_{j,k})
\end{equation}
where $p_{j,k}$ represents the probability that patch $\textbf{x}_{j}$ belongs to class $k$.

The patches with higher information entropy are considered as valuable training samples, hence we treat them as preferred candidates. The patches in ${\textbf{U}}_{1}$ are then sorted according to the descending order of $E_{j}$ value, forming a sample set ${\textbf{U}}_{2}={\{{\hat{\textbf{x}}}_{j}\}}^{J}_{j=1}$. Considering that data with low information entropy provides insufficient information, we only use the top $\mu$ candidates of each category in the set ${\textbf{U}}_{2}$ to perform retrieval as follows.

Given a patch ${\hat{\textbf{x}}}_{j}$ that possesses the pseudo-label ${\hat{l}}_{j}$, we take it as a query image and retrieve its similar samples from the source domain $\textbf{\emph{D}}_{\textbf{\emph{S}}}$. We use the deep features extracted from full-connected layer of the pre-trained ResNet-50 for retrieval. The similarities between ${\hat{\textbf{x}}}_{j}$ and the source domain samples are measured by the Euclidean distance.

Then, we use the existing labels of the source domain to determine the confidence of the pseudo-labels. If the top $\delta$ retrieved results from the source domain have the same label $g$, and $g$ is the same as the pseudo-label ${\hat{l}}_{j}$ of the query patch ${\hat{\textbf{x}}}_{j}$ ({\em i.e.} ${\hat{l}}_{j}=g$), ${\hat{\textbf{x}}}_{j}$ is considered to be a relevant sample. Otherwise, ${\hat{\textbf{x}}}_{j}$ is removed from the set ${\textbf{U}}_{2}$. After sample screening, the remainders of ${\textbf{U}}_{2}$ form a new target domain set ${\textbf{U}}_{tg}$. Finally, the image patches along with their corresponding pseudo-labels in the set ${\textbf{U}}_{tg}$ are used to fine-tune a CNN model that is specific to the target image.

\subsection{A hybrid land-cover classification algorithm}
\label{sec:classification}

\begin{figure*}[b!]
\centering
\includegraphics[width=1\textwidth]
{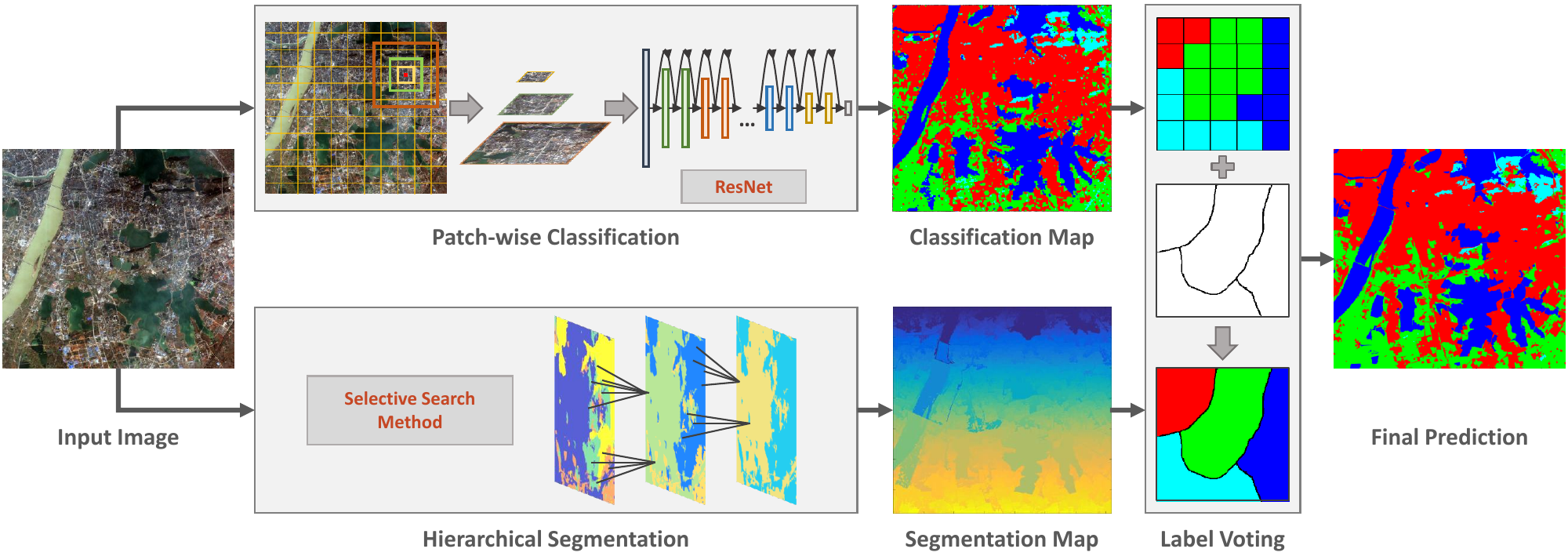}
\caption{The proposed land-cover classification approach.}
\label{figure:classification}
\end{figure*}

Land-cover classification aims to assign pixels in a RS image with land-cover category labels. Both the category and boundary information of the ground objects is essential for accurate classification. We therefore propose a hybrid algorithm, which combines patch-wise classification and hierarchical segmentation through a majority voting strategy, as shown in Fig. \ref{figure:classification}.

\subsubsection{Patch-wise classification}
\begin{figure*}[htb!]
\centering
\includegraphics[width=0.85\textwidth]
{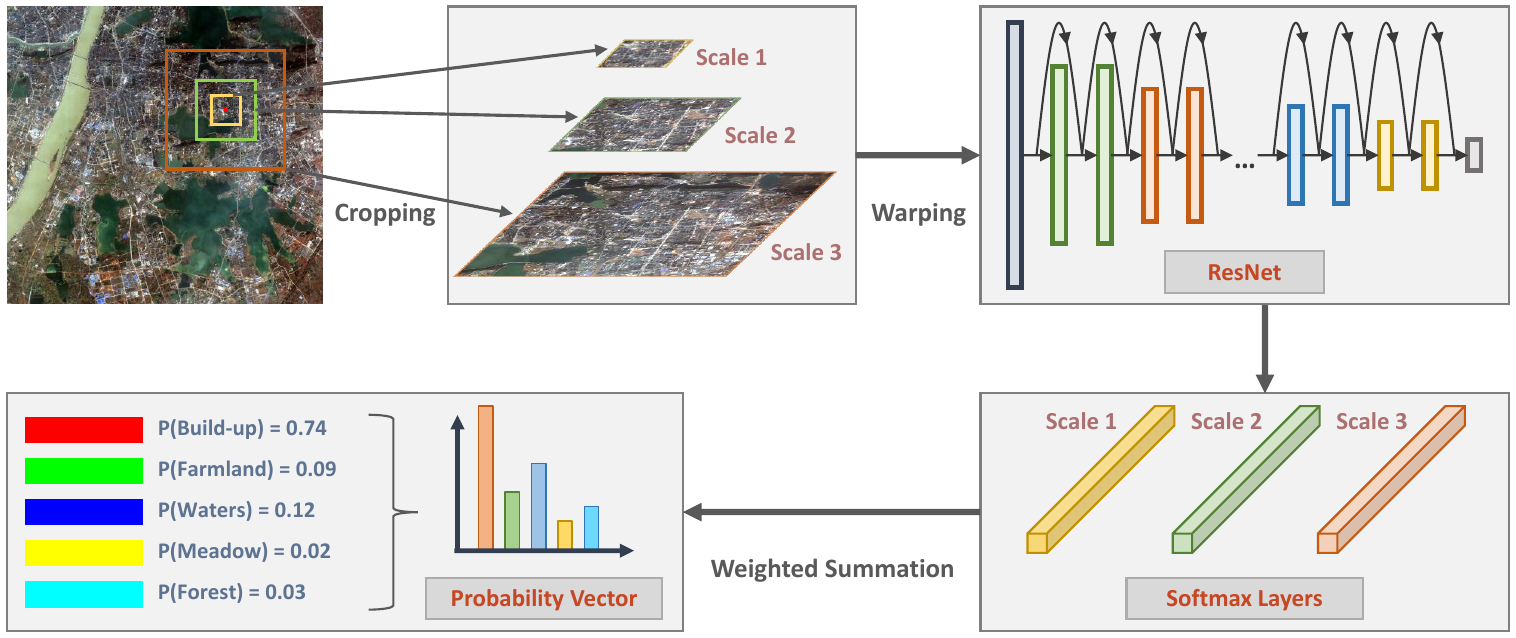}
\caption{Multi-scale contextual information aggregation.}
\label{figure:multi-scale}
\end{figure*}

Since the ground objects show different characteristics in different spatial resolutions, it is difficult to capture sufficient information of objects from the single-scale observation field. To exploit the attributes of the objects and their spatial distributions, we propose to utilize multi-scale contextual information for classification, which is illustrated in Fig. \ref{figure:multi-scale}.

The target image $\chi_{\textbf{\emph{T}}}$ is partitioned into non-overlapping patches $\textbf{U}={\{\textbf{x}_{i}\}}^{I}_{i=1}$ by grid with the size of $s_{1}\times s_{1}$ pixels ($s_{1}$ is the minimum value in the succession of scales). For each patch $\textbf{x}_{i}$, its center pixel is regarded as a reference pixel $\textbf{z}$. Around $\textbf{z}$, a series of patches with sizes of $s_{2}\times s_{2}, \ldots, s_{N}\times s_{N}$ pixels are sampled, so that each reference pixel possesses $N$ multi-scale samples. Then, these multi-scale patches are uniformly resized to $224\times 224$ and are input to the ResNet-50 model. After forward propagation, the classification probability vector ${\textbf{p}}_{s_{n}}(\textbf{z})$ of scale $s_{n}$ at pixel $\textbf{z}$ is obtained from softmax layer:

\begin{equation}
{\textbf{p}}_{s_{n}}(\textbf{z}) = \{{p}_{s_{n},1}(\textbf{z}),{p}_{s_{n},2}(\textbf{z}),\ldots,{p}_{s_{n},K}(\textbf{z})\}, {\textbf{p}}_{s_{n}}(\textbf{z})\in{\mathbb{R}}^{K}
\end{equation}
where $n\in\{1,\ldots,N\}$, ${p}_{s_{n},k}(\textbf{z})$ represents the probability that $\textbf{z}$ belongs to class $k$ at the $n$-th scale.

Contextual information of multi-scale patches is aggregated using a weighted fusion strategy. The specificity measure \cite{weight}, which describes the certainty of classification result, is employed as the weight:

\begin{equation}
{W}_{s_{n}}(\textbf{z}) = \sum_{k=1}^{K-1}\\\frac{1}{k}\cdot({\hat{p}}_{s_{n},k}(\textbf{z})-{\hat{p}}_{s_{n},k+1}(\textbf{z}))
\end{equation}
where $\{{\hat{p}}_{s_{n},1}(\textbf{z}),{\hat{p}}_{s_{n},2}(\textbf{z}),\ldots,{\hat{p}}_{s_{n},K}(\textbf{z})\}$ is the descending order of the vector ${\textbf{p}}_{s_{n}}(\textbf{z})$. The value of ${W}_{s_{n}}(\textbf{z})$ ranges from 0 to 1, and the higher value signifies the higher categorization confidence. The weighted probability ${\tilde{\textbf{p}}}_{k}(\textbf{z})$ is expressed as:

\begin{equation}
{\tilde{\textbf{p}}}_{k}(\textbf{z})=\frac{\sum_{n=1}^{N}\\{W}_{s_{n}}(\textbf{z})\cdot {p}_{s_{n},k}(\textbf{z})}{\sum_{n=1}^{N}\\{W}_{s_{n}}(\textbf{z})}
\end{equation}

where ${\tilde{\textbf{p}}}_{k}(\textbf{z})\in[0,1]$ indicates the probability that the reference pixel $\textbf{z}$ belongs to class $k$. The aggregated probabilities of all categories can constitute a new classification probability vector. Then the reference pixel $\textbf{z}$ is classified as:

\begin{equation}
l(\textbf{z}) = \mathop{\arg\max}_{k\in\{1,\ldots,K\}}\\{\tilde{\textbf{p}}}_{k}(\textbf{z})
\end{equation}
where $l(\textbf{z})$ is the category label of the pixel $\textbf{z}$. Then, we assign the label $l(\textbf{z})$ to each pixel in the patch $\textbf{x}_{i}$. After classifying all the patches in the entire RS image, a patch-wise classification map $\chi_{\textbf{\emph{c}}}$ is therefore acquired.

\subsubsection{Object-based voting}
To obtain precise boundary information of the objects, we utilize segmentation map generated by selective search method \cite{selectiveSearch} to refine the preliminary classification map. Selective search is a hierarchical segmentation method. It exploits a graph-based approach \cite{graphSegmentation} to produce a series of initial regions in different color spaces, and then uses the greedy algorithm to iteratively merge small regions. The color, texture, size and fill similarities are employed to control the merging level. Since various image features are considered in the process of initial segmentation and iterative merging, selective search can produce high-quality segmentation results.

After obtaining classification and segmentation maps by patch-wise classification and selective search, we integrate the category and boundary information using a majority voting strategy. Let $\textbf{V}={\{\textbf{y}_{f}\}}^{F}_{f=1}$ denote the homogeneous regions in the segmentation map $\chi_{\textbf{\emph{s}}}$ generated from the target image $\chi_{\textbf{\emph{T}}}$. And ${\hat{\textbf{y}}}_{f}$ is the corresponding area of $\textbf{y}_{f}$ in the classification map $\chi_{\textbf{\emph{c}}}$. The number of pixels contained by ${\hat{\textbf{y}}}_{f}$ is $M=|\textbf{y}_{f}|$, and category label of the $m$-th pixel is ${l}_{m}$, $m\in\{1,\ldots,M\}$. Then the number of pixels belonging to each class in ${\hat{\textbf{y}}}_{f}$ is counted, and the most frequent label $T(\textbf{y}_{f})$ is assigned to all pixels in $\textbf{y}_{f}$:

\begin{equation}
T(\textbf{y}_{f}) = \mathop{\arg\max}_{r\in\{1,\ldots,K\}}\\\sum_{m=1}^{M}\\sign(l_{m}=r)
\end{equation}
where $sign(\cdot)$ is an indicator function, $sign(true)=1$, $sign(false)=0$, and $r$ denotes the possible class label. For all segmented objects, the same voting scheme is applied, and the final classification result is then acquired.

\section{GID: a well-annotated dataset for land-cover classification}
\label{sec:data}
\begin{figure*}[htb!]
\centering
\includegraphics[width=1\linewidth]
{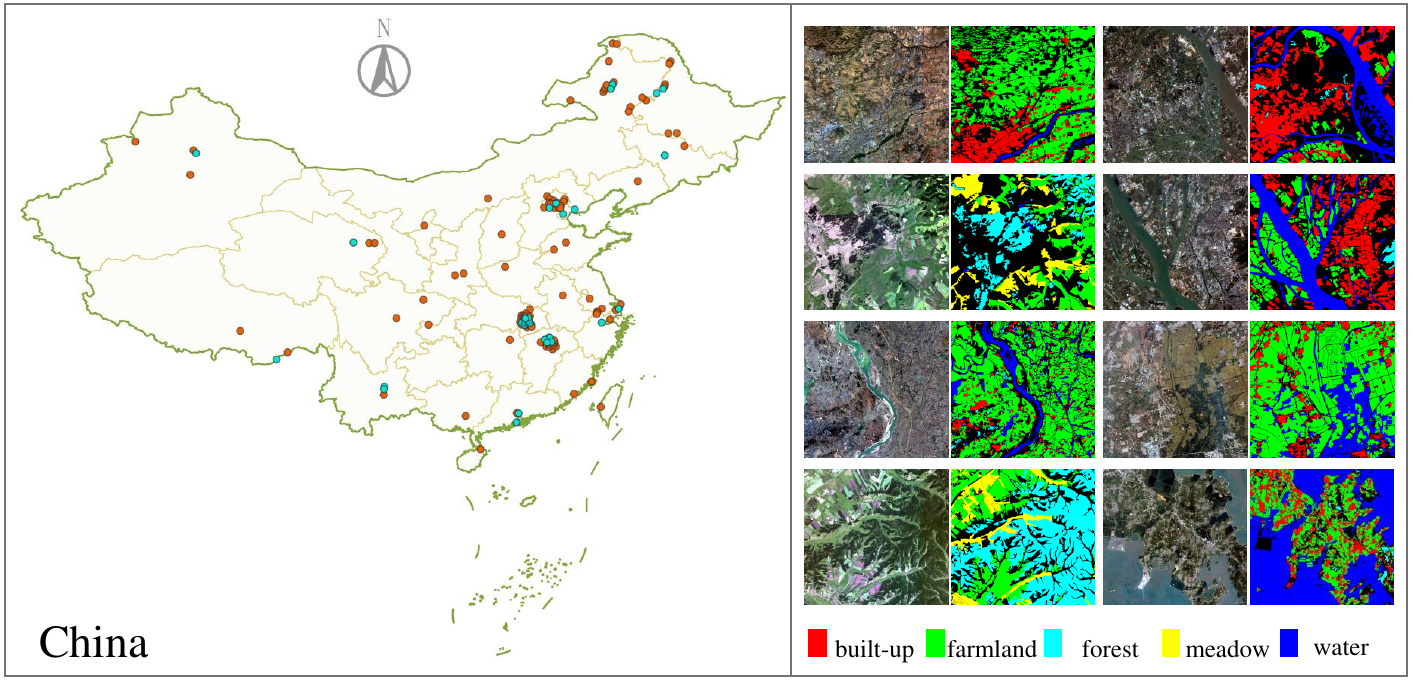}
\caption{Left: Distribution of the geographical locations of images in GID. The large-scale classification set contains 120 training images and 30 validation images, which are marked with orange and cyan. Right: Examples of GF-2 images and their corresponding ground truth.}
\label{figure:samplesGIDnew}
\end{figure*}

We construct a large-scale land-cover dataset with Gaofen-2 (GF-2) satellite images. This new dataset, which is named as Gaofen Image Dataset (GID), has superiorities over the existing land-cover dataset because of its large coverage, wide distribution, and high spatial resolution. GID consists of two parts: a large-scale classification set and a fine land-cover classification set. The large-scale classification set contains 150 pixel-level annotated GF-2 images (see Fig. \ref{figure:samplesGIDnew}), and the fine classification set is composed of 30,000 multi-scale image patches (see Fig. \ref{figure:samplesGID15}) coupled with 10 pixel-level annotated GF-2 images. More details are shown in Table \ref{table:O}. The training and validation data with 15 categories is collected and re-labeled based on the training and validation images with 5 categories, respectively.

\begin{table*}[htb!]
\caption{Composition of GID dataset.}
\small
\arrayrulecolor{setblue}
\renewcommand\arraystretch{1.1}
\resizebox{\textwidth}{!}{
\begin{tabular}{c|c|c|c|c}
\toprule[1.2pt]
\rowcolor{setgray}
Set&Class&Training&Size&Validation\\
\midrule[0.6pt]
Large-scale Classification&5&120 GF-2 images&$6800\times 7200$&30 GF-2 images\\
\midrule[0.6pt]
Fine Land-cover Classification&15&30,000 patches&$56\times 56$, $112\times 112$, $224\times 224$&10 GF-2 images\\
\bottomrule[1.2pt]
\end{tabular}}
\label{table:O}
\end{table*}

The training images of GID are utilized to pre-train a CNN model with strong generalization ability specific to RS domain. In addition, it can serve as data resource to advance the state-of-the-art in land-cover classification task. GID and its reference annotations have been provided online at \url{https://x-ytong.github.io/project/GID.html}.

Furthermore, to validate the transferability of our method on multi-source HRRS images, we annotate high-resolution images acquired by different sensors, including Gaofen-1, Jilin-1, Ziyuan-3, Sentinel-2A satellite images, and Google Earth platform data. GID and multi-source images are introduced in Section \ref{sec:GID} and Section \ref{sec:multi-source data}, respectively.

\subsection{Gaofen image dataset}
\label{sec:GID}

\subsubsection{Gaofen-2 satellite images}
Gaofen-2 (GF-2) is the second satellite of ¡°High-definition Earth Observation System (HDEOS)¡± promoted by China National Space Administration (CNSA). Two panchromatic and multispectral (PMS) sensors with effective spatial resolution of 1 $m$ panchromatic (pan)/4 $m$ multispectral (MS) are onboard the GF-2 satellite, with a combined swath of 45 $km$. The resolution of the sub-satellite point is 0.8 $m$ pan/3.24 $m$ MS, and the viewing angle of a single camera is $2.1^\circ$. GF-2 satellite realizes global observation within 69 days and repeat observations within 5 days.

The multispectral image we used to establish GID provide a spectral range of blue (0.45-0.52 $\mu m$), green (0.52-0.59 $\mu m$), red (0.63-0.69 $\mu m$) and near-infrared (0.77-0.89 $\mu m$), and a spatial dimension of $6800\times 7200$ pixels covering a geographic area of 506 $km^{2}$.

GF-2 images achieve a combination of high spatial resolution and wide field of view, allowing the observation of detailed information over large areas. Since launched in 2014, GF-2 has been made use of for land-cover surveys, environmental monitoring, crop estimation, construction planning and other important applications.

\begin{figure*}[htb!]
\centering
\includegraphics[width=0.95\linewidth]
{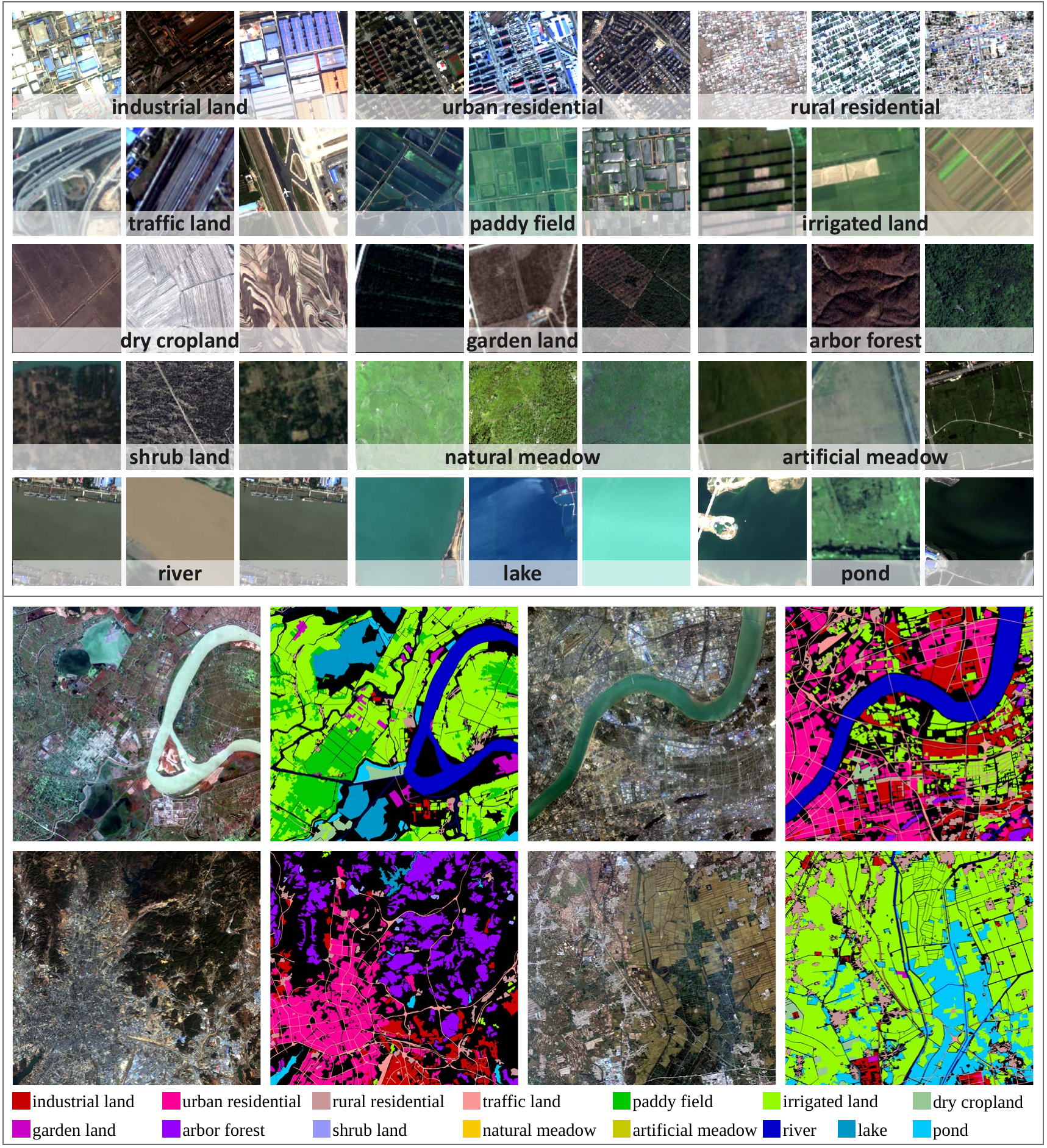}
\caption{Top: Training samples in the fine land-cover classification set. Three examples of each category are shown. There are 30,000 images within 15 classes. Bottom: Examples of validation images and their corresponding ground truth.}
\label{figure:samplesGID15}
\end{figure*}

\subsubsection{Land-cover types}
\begin{figure*}[htb!]
\centering
\includegraphics[width=0.8\linewidth]
{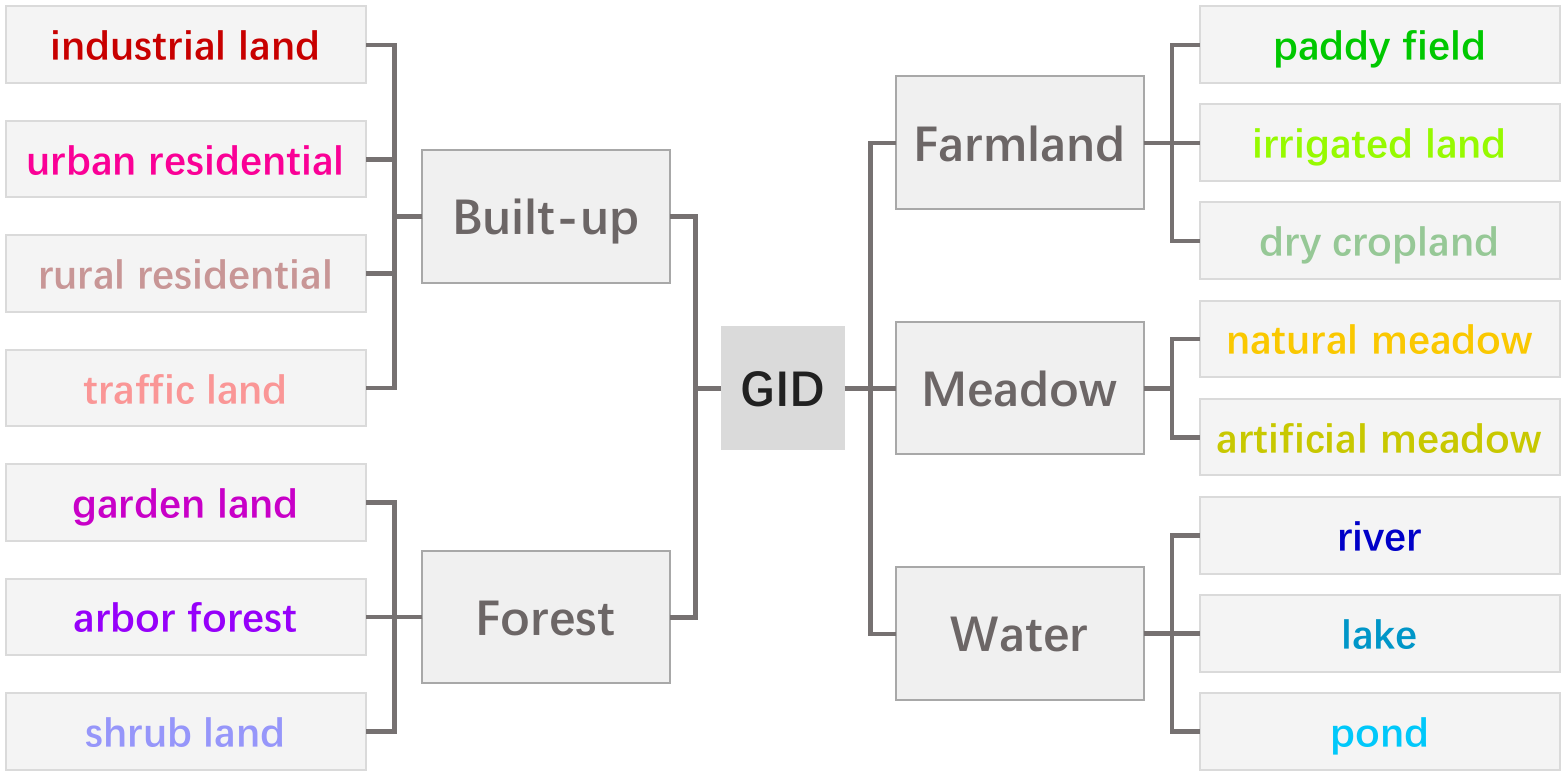}
\caption{Classification criteria for GID dataset}
\label{figure:hiercalclasses}
\end{figure*}

We refer to Chinese Land Use Classification Criteria (GB/T21010-2017) to determine a hierarchical category system. In the large-scale classification set of GID, 5 major categories are annotated: \emph{built-up}, \emph{farmland}, \emph{forest}, \emph{meadow}, and \emph{water}, which are pixel-level labeled with five different colors: red, green, cyan, yellow, and blue, respectively. Areas not belonging to the above five categories and clutter regions are labeled as background, which is represented using black color. The fine land-cover classification set is made up of 15 sub-categories: \emph{paddy field}, \emph{irrigated land}, \emph{dry cropland}, \emph{garden land}, \emph{arbor forest}, \emph{shrub land}, \emph{natural meadow}, \emph{artificial meadow}, \emph{industrial land}, \emph{urban residential}, \emph{rural residential}, \emph{traffic land}, \emph{river}, \emph{lake}, and \emph{pond}. Its training set contains 2,000 patches per class, and validation images are labeled in pixel level. Examples of training and validation data with 15 categories are demonstrated in Fig. \ref{figure:samplesGID15}. The two parts of GID constitute a hierarchical classification system, and the affiliation of them is shown in Fig. \ref{figure:hiercalclasses}.

\subsubsection{Dataset properties}

\textbf{Widely distributed:} GID contains 150 high-quality GF-2 images acquired from more than 60 different cities in China, which is shown in Fig. \ref{figure:samplesGIDnew}. It is widely distributed over the geographic areas covering more than 50,000 $km^{2}$. Due to the extensive geographical distribution, GID represents the distribution information of ground objects in different areas.

\begin{figure*}[!htb]
\centering
\includegraphics[width=1\textwidth]
{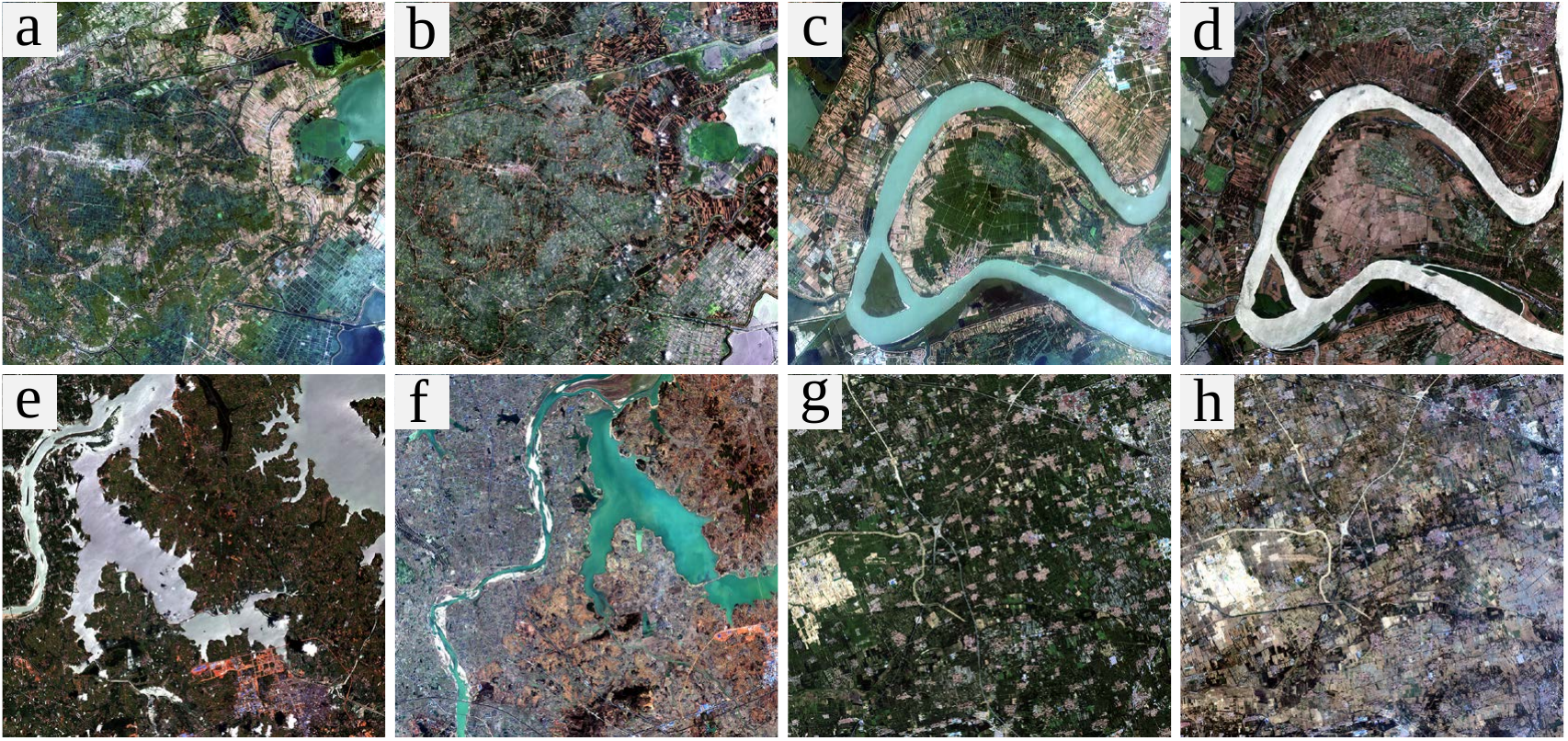}
\caption{Multi-temporal images captured from the same locations or overlapping areas.}
\label{figure:multitemporal}
\end{figure*}

\textbf{Multi-temporal:} The images obtained at different time from the same location or overlapping areas are included in GID. For example, Fig. \ref{figure:multitemporal}(a)-(b) are images acquired at Xiantao, Hubei Province on September 2, 2015 and June 14, 2016 respectively. Fig. \ref{figure:multitemporal}(c)-(d) are images captured at Wuhan, Hubei Province on September 2, 2015 and June 14, 2016. Fig. \ref{figure:multitemporal}(e)-(f) are images acquired around Nanchang, Jiangxi Province on August 12, 2016 and January 3, 2015. Fig. \ref{figure:multitemporal}(g)-(h) are images acquired around Langfang, Hebei Province on August 27, 2016 and June 9, 2016. The spectral responses of the ground objects in the same area emerge in distinct differences due to seasonal changes.

Therefore, GID presents rich diversity of the ground objects in spectral response and morphological structure.

\subsection{Gaofen-1, Jilin-1, Ziyuan-3, Sentinel-2A images, and Google Earth data}
\label{sec:multi-source data}
Newly acquired RS images may come from different sensors, the classification of the multi-source images is therefore of great significance. We also verify the effectiveness of our algorithm on HRRS images captured by other sensors, including Gaofen-1 (GF-1), Jilin-1 (JL-1), Ziyuan-3 (ZY-3), Sentinel-2A (ST-2A), and Google Earth platform data. The introduction of these images is as follows.

\begin{figure*}[!htb]
\centering
\includegraphics[width=0.85\linewidth]
{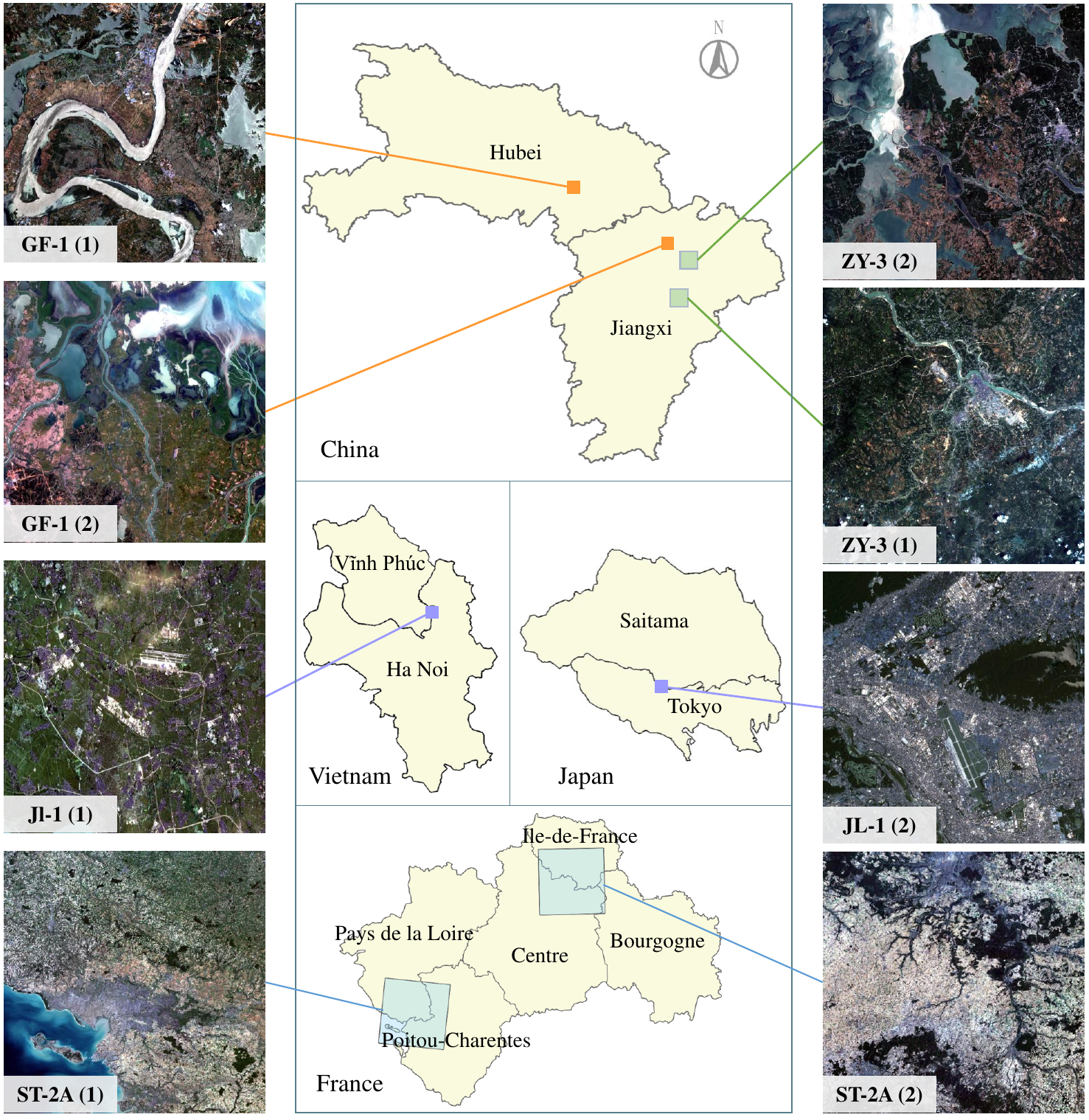}
\caption{GF-1, JL-1, ZY-3, ST-2A satellite images and their acquisition locations.}
\label{figure:samplesMultinew}
\end{figure*}

\textbf{Gaofen-1 Satellite Images:} GF-1 satellite configures with two PMS and four wide field of view (WFV) sensors. The resolution of PMS is 2 m pan/8 m MS, and the swath is 60 km. Two GF-1 multispectral images that were captured in Wuhan, Hubei Province on July 25, 2016, and in Jiujiang, Jiangxi Province on October 16, 2015 are employed in the experiments. We denote them as GF-1(1) and GF-1(2), as shown in Fig. \ref{figure:samplesMultinew}.

\textbf{Jilin-1 Satellite Images:} The resolution of JL-1 satellite is 0.72 m pan/2.88 m MS, and it has only three bands of red, green, and blue. Two JL-1 images that were respectively captured around Ha Noi, Vietnam on June 11, 2016, and around Tokyo, Japan on June 3, 2016 are used in the experiments. We denote them as JL-1(1) and JL-1(2), as shown in Fig. \ref{figure:samplesMultinew}.

\textbf{Ziyuan-3 Satellite Images:} ZY-3 satellite configures with three panchromatic time delay integration (TDI) charge coupled device (CCD) sensors and a multispectral scanner (MSS). The resolution of MSS is 5.8 m, and the swath is 52 km. Two ZY-3 multispectral images that were respectively captured in Fuzhou, Jiangxi Province on August 28, 2016, and in Shangrao, Jiangxi Province on August 28, 2016 are utilized in the experiments. We denote them as ZY-3 (1) and ZY-3 (2), as shown in Fig. \ref{figure:samplesMultinew}.

\textbf{Sentinel-2A Satellite Images:} ST-2A satellite carries a PMS sensor that covers 13 spectral bands. The spatial resolution of visible and near-infrared bands is 10m. ST-2A satellite has a very wide swath of 290 km. Two ST-2A multispectral images that were respectively captured around La Rochelle, France on October 21, 2018, and around Orleans, France on October 18, 2018 are used in the experiments. We denote them as ST-2A(1) and ST-2A(2), as shown in Fig. \ref{figure:samplesMultinew}.

\begin{figure*}[htb!]
\centering
\includegraphics[width=0.6\textwidth]
{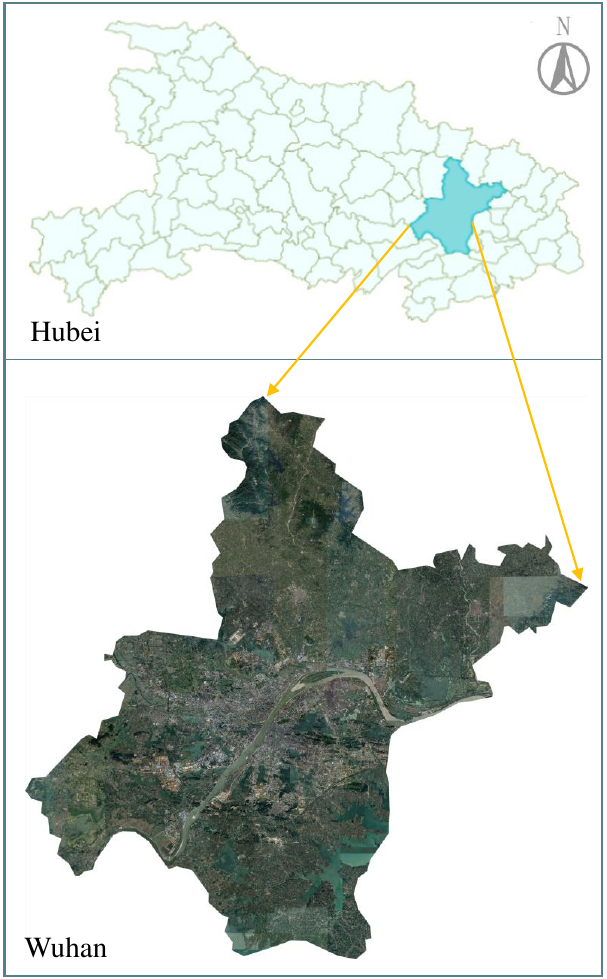}
\caption{Google Earth platform data in Wuhan, Hubei province, China.}
\label{figure:samplesGEWH}
\end{figure*}

\textbf{Google Earth Platform Data:} To confirm the practicality of our algorithm for application, we conduct land-cover classification in Wuhan, Hubei province, China. Google Earth platform data captured on December 9, 2017 from Wuhan area are utilized. They have the resolution of 4.78 m and contains only three bands of red, green, and blue. We refer to these images as GE-WH, as shown in Fig. \ref{figure:samplesGEWH}.

\begin{figure*}[htb!]
\centering
\includegraphics[width=1\linewidth]
{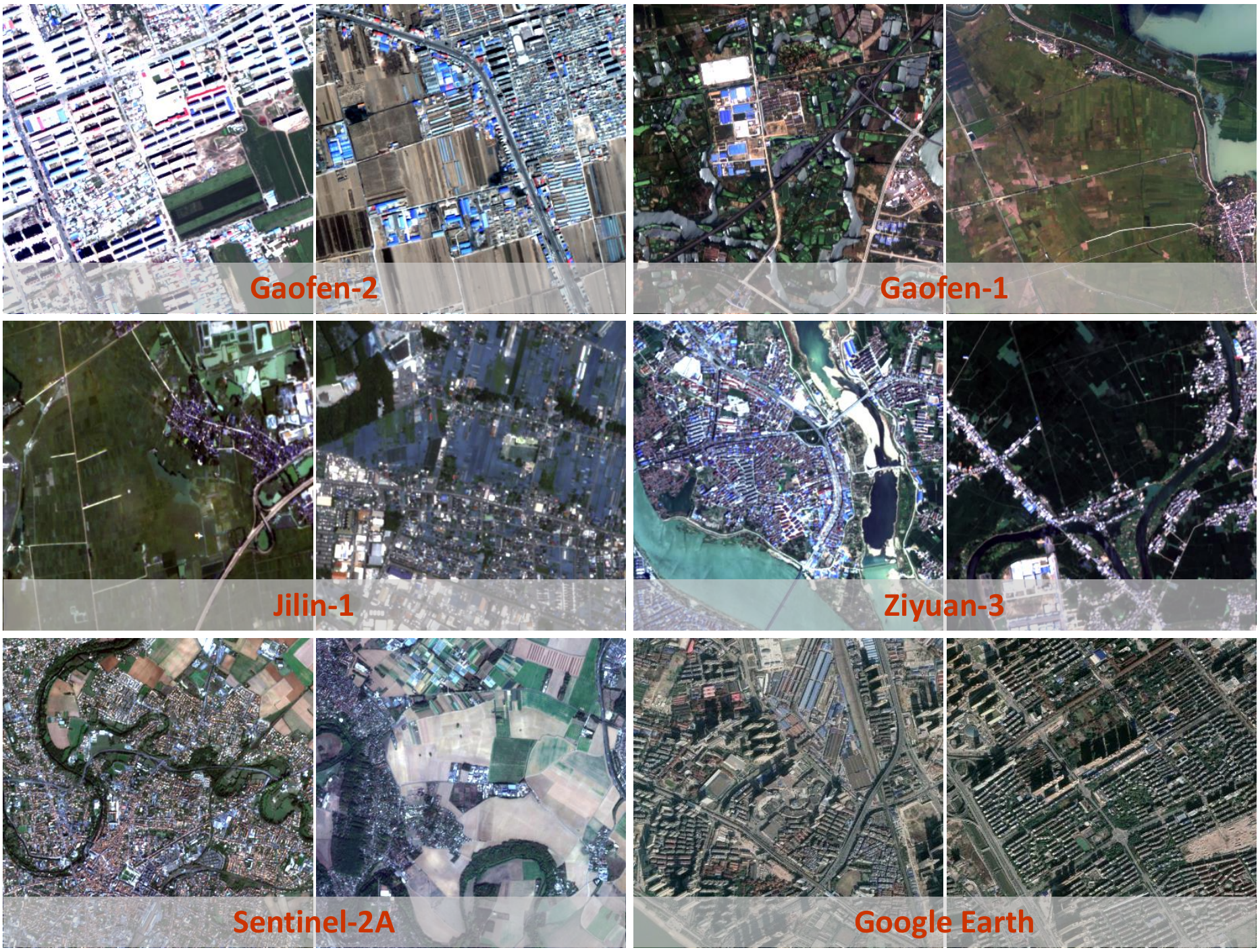}
\caption{The heterogeneity of multi-source data.}
\label{figure:heterogeneity}
\end{figure*}

We demonstrate the heterogeneity of multi-source data in Fig. \ref{figure:heterogeneity}. Sub images with size of $400\times 400$ are cropped from 6 different data. It can be observed that, besides the significant difference in spatial resolution, the morphology of \emph{farmland} and \emph{built-up} varies greatly due to the diversity of geographic location. In addition, because of seasonal changes, part of \emph{farmland} is fallow and the other part is covered by crops. The heterogeneity of the source and target domains brings challenges to transfer learning.

\section{Experimental results}
\label{sec:experiments}
We test our algorithm and analyse the experimental results in this section. Two types of land-cover classification issues are examined: 1) transferring deep models to classify HRRS images captured with the same sensor and under different conditions, 2) transferring deep models to classify multi-source HRRS images. For performance comparison, several object-based land-cover classification methods are utilized. The implementation details, comparison methods, and evaluation metrics are introduced in Section \ref{sec:setup}. Section \ref{sec:Gaofen-2 results} presents the experimental results of Gaofen-2 (GF-2) images. Section \ref{sec:multi-source results} tests the transferability of our algorithm on multi-source images.

\subsection{Experimental setup}
\label{sec:setup}

\textbf{Pre-processing:} For pre-processing, we re-quantize the responses of GF-2, GF-1, JL-1, ST-2A, and ZY-3 images to 8-bit with the optimized linear stretch function embedded in ENVI software. For GE-WH, we perform no pre-processing. In particular, to classify 3-band images (JL-1, GE-WH), we remove the near-infrared band of GF-2 images to train the model with the input size of 3 channels. When annotating label masks, we use the lasso tool in Adobe Photoshop to mark the areas of each land-cover category in the images.

\textbf{Model Training:} ResNet-50 models are pre-trained on the training images of GID, and our algorithm is tested on the validation images of GID as well as multi-source images. For the large-scale classification set, we train the models using patches with multiple scales to exploit the multi-scale contextual information. Patches of size $56\times 56$, $112\times 112$, and $224\times 224$ are randomly sampled on each training image. If more than 80\% pixels in a patch are covered by the same category, this patch is considered as a training sample. For each of the 5 categories, 10,000 samples are selected for each scale. Thus, a total of $10,000\times 3\times 5 = 150,000$ patches are collected. Then, they are uniformly resized to $224\times 224$ to pre-train a ResNet-50 model. For the fine land-cover classification set, we directly use the 30,000 image patches to fine-tune the ResNet-50 model pre-trained on 5 categories. In addition, image augmentation strategies \cite{AlexNet} are adopted to avoid overfitting.

For multi-source images, we separately partition them into candidate patch sets with multi-scale sliding windows. We set different window sizes for different data according to their spatial resolution. For GF-1, the window sizes are $28\times 28$, $56\times 56$, and $112\times 112$. For JL-1, the window sizes are $78\times 78$, $156\times 156$, and $312\times 312$. For ZY-3, the window sizes are $38\times 38$, $76\times 76$, and $152\times 152$. For ST-2A, the window sizes are $22\times 22$, $44\times 44$, and $88\times 88$. For GE-WH, the window sizes are $46\times 46$, $92\times 92$, and $184\times 184$. And the length of the stride is set as half the size of the window.

The parameters of ResNet-50 are initialized with ImageNet \cite{ImageNET}, and the softmax layer is initialized by Gaussian distribution. The last three bottlenecks and softmax layer of ResNet-50 are trained. In particular, when pre-training ResNet-50 for 4-band data, the new conv1 layer is initialized by Gaussian distribution, and is trained along with the last three bottlenecks and softmax layer. To train a deep model with multi-scale patches, patches with multiple scales are uniformly resized to $224\times 224$. Considering the differences in feature distribution between multi-source images, we select $\textbf{U}_{tg}$ set and fine-tune ResNet-50 model for each target image separately to prevent different data sources from interfering with each other. The hyper-parameters for training are set as follows: batch size is 32, epoch number is 15, momentum value is 0.9, and initial learning rate is 0.1. When the error rate stops decreasing, we divide the learning rate by 10 and use the new value to update the parameters.

\textbf{Comparison Methods:} We compare our algorithm with several object-based classification methods. Selective search method is used to segment the image into homogeneous objects. Specifically, a set of four different features are exploited, including spectral feature, gray-level co-occurrence matrix (GLCM) \cite{GLCM}, differential morphological profiles (DMP) \cite{DMP}, and local binary patterns (LBP) \cite{LBP}. Moreover, we consider multi-feature fusion strategy, which aggregates the above features by normalization and vector concatenation. Maximum likelihood classification (MLC), random forest (RF), support vector machine (SVM), and multi-layer perceptron (MLP) are employed as classifiers.

The parameters of the comparison methods are set to the optimal values. The window size is set to $7\times 7$ pixels for GLCM. The radius of the structural elements for DMP is set to 4 pixels. And for LBP, the filter size is $5\times 5$ pixels. The number of trees for RF is 500. The kernel function of SVM is radial basis function (RBF) kernel. MLP has 4 hidden layers with 20 nodes per layer. The initial segmentation size is set to 400 for selective search. We randomly select 15,000 multi-scale patches from GID's training set to train comparison classifiers. After training, the classifiers are directly used to classify the target data.

\textbf{Evaluation Metrics:} To evaluate our algorithm, we assess the experimental results with Kappa coefficient (Kappa), overall accuracy (OA), and user's accuracy \cite{accuracy}.

We test our algorithm on the validation set of GID and the multi-source data. The classification accuracy is assessed on all of the labeled pixels (except for the background) in the test images. Let $P_{ab}$ denote the number of pixels of class $a$ predicted to belong to class $b$, and let $t_{a}=\sum_{b}P_{ab}$ be the total number of pixels belong to class $a$, let $t_{b}=\sum_{a}P_{ab}$ be the total number of pixels predicted to class $b$. The metrics are defined as follow:
\begin{itemize}
\item[-]\emph{Kappa coefficient}: Kappa is a statistic that measures the agreement between the prediction and the ground truth.

\begin{equation}
Kappa = \frac{P_{o}-P_{c}}{1-P_{c}}
\end{equation}
where

\begin{equation}
P_{o} = \frac{\sum_{a}\\P_{aa}}{\sum_{a}\\t_{a}}
\end{equation}

\begin{equation}
P_{c} = \frac{\sum_{k}\\(\sum_{b}P_{kb}\cdot\sum_{a}P_{ak})}{\sum_{a}t_{a}\cdot\sum_{a}t_{a}}
\end{equation}
where $k\in[1,K]$, and $K$ is the number of categories.

\item[-]\emph{Overall accuracy}: OA is the percentage of correctly classified pixels and all pixels in the entire image.

\begin{equation}
OA = \frac{\sum_{a}\\P_{aa}}{\sum_{a}\\t_{a}}
\end{equation}

\item[-]\emph{User's accuracy}: User's accuracy of class $b$ is the proportion of correctly classified pixels in all pixels predicted to class $b$.

\begin{equation}
user's\ accuracy = \frac{P_{bb}}{t_{b}}
\end{equation}
\end{itemize}

The values of Kappa, OA, and user's accuracy are in the range of 0 to 1, and the higher value indicates the better classification performance.

\subsection{Experiments on Gaofen-2 images}
\label{sec:Gaofen-2 results}

\begin{table*}[b!]
\caption{Comparison of different land-cover classification methods on GID.}
\arrayrulecolor{setblue}
\renewcommand\arraystretch{1.1}
\resizebox{\textwidth}{!}{
\begin{tabular}{cccccccccccc}
\toprule[1.2pt]
\rowcolor{setgray}&\multicolumn{7}{>{\columncolor{setgray}}c}{\textbf{5 Classes}}&&\multicolumn{2}{>{\columncolor{setgray}}c}{\textbf{15 Classes}}\\
\cmidrule[0.6pt]{2-8}\cmidrule[0.6pt]{10-11}
\rowcolor{setgray}

&&&\multicolumn{5}{>{\columncolor{setgray}}c}{User's Accuracy (\%)}&&&\\
\cmidrule[0.6pt]{4-8}
\rowcolor{setgray}
\multirow{-3}{*}{\textbf{Methods}}&\multirow{-2}{*}\textbf{Kappa}&\multirow{-2}{*}\textbf{OA(\%)}
&built-up&farmland&forest&meadow&water&&
\multirow{-2}{*}\textbf{Kappa}&\multirow{-2}{*}\textbf{OA(\%)}\\
\midrule[0.6pt]
MLC+spectral&0.504&65.48&58.28&66.65&33.25&4.23&82.30&&0.134&22.65\\
MLC+GLCM&0.389&56.22&66.65&73.56&20.55&0.05&63.50&&0.092&16.07\\
MLC+DMP&0.512&65.82&59.05&68.04&33.91&4.19&82.53&&0.028&23.26\\
MLC+LBP&0.195&37.31&39.77&52.39&25.08&0.96&66.82&&0.084&16.28\\
MLC+Fusion&0.606&74.53&61.81&67.38&36.15&2.92&82.09&&0.145&23.61\\
\cmidrule[0.6pt]{1-1}
RF+spectral&0.526&68.73&55.48&67.25&34.76&2.22&84.25&&0.164&23.79\\
RF+GLCM&0.426&61.83&65.73&70.71&18.84&0.68&67.54&&0.119&19.05\\
RF+DMP&0.512&67.04&56.38&67.08&34.97&3.22&83.40&&0.173&24.52\\
RF+LBP&0.365&58.76&46.14&59.42&23.19&1.40&68.30&&0.063&11.60\\
RF+Fusion&0.641&78.45&62.61&71.12&36.10&3.94&84.29&&0.237&33.70\\
\cmidrule[0.6pt]{1-1}
SVM+spectral&0.103&46.11&54.68&42.86&41.82&1.02&62.11&&0.024&22.72\\
SVM+GLCM&0.456&62.15&72.67&68.64&20.41&0.91&67.13&&0.096&16.65\\
SVM+DMP&0.125&47.12&54.14&44.47&40.78&0.13&70.97&&0.130&19.01\\
SVM+LBP&0.293&51.94&44.65&59.02&20.40&1.89&45.83&&0.027&12.28\\
SVM+Fusion&0.488&66.88&61.28&72.27&23.01&2.26&54.18&&0.148&23.92\\
\cmidrule[0.6pt]{1-1}
MLP+spectral&0.442&60.93&52.42&58.26&29.21&1.36&84.11&&0.082&14.19\\
MLP+GLCM&0.440&61.21&74.31&68.94&19.95&1.00&66.48&&0.082&16.65\\
MLP+DMP&0.480&63.05&55.81&65.25&41.38&1.55&82.53&&0.162&26.06\\
MLP+LBP&0.220&41.30&38.90&49.75&15.70&0.82&61.03&&0.104&18.63\\
MLP+Fusion&0.616&75.81&58.69&72.40&32.86&2.67&83.99&&0.199&30.57\\
\midrule[0.6pt]
PT-GID&\textbf{0.924}&\textbf{96.28}&\textbf{88.42}&\textbf{91.85}&\textbf{79.42}&\textbf{70.55}&\textbf{87.60}&&\textbf{0.605}&\textbf{70.04}\\
\bottomrule[1.2pt]
\end{tabular}}
\label{table:onGID}
\end{table*}

To test the effectiveness of CNN model pre-trained on GID, which is denoted as PT-GID, we compare our approach with object-based methods. Although the acquisition location and time are diverse, the objects belong to the same class have similar spectral response in the images captured by the same sensor ({\em i.e.} GF-2 satellite). Hence we directly use ResNet-50 pre-trained on GID's training set to classify the validation images. Multi-scale information aggregation is exploited here, and the initial segmentation size is 400 pixels. The experimental results of our algorithm and the comparison methods are shown in Table \ref{table:onGID}.

It can be seen that PT-GID achieves the highest Kappa and OA of 0.924 and 96.28\% on 5 classes, and of 0.605 and 70.04\% on 15 classes. This experimental phenomenon shows the strong transferability of PT-GID. However, among the comparison methods, the optimal results are given by RF+Fusion, yielding Kappa and OA of 0.641 and 78.45\% on 5 classes, and of 0.237 and 33.70\% on 15 classes, respectively. This shows that conventional classifiers and features lack generalization capabilities for data shift.

\begin{figure*}[tb!]
\centering
\includegraphics[width=1\textwidth]
{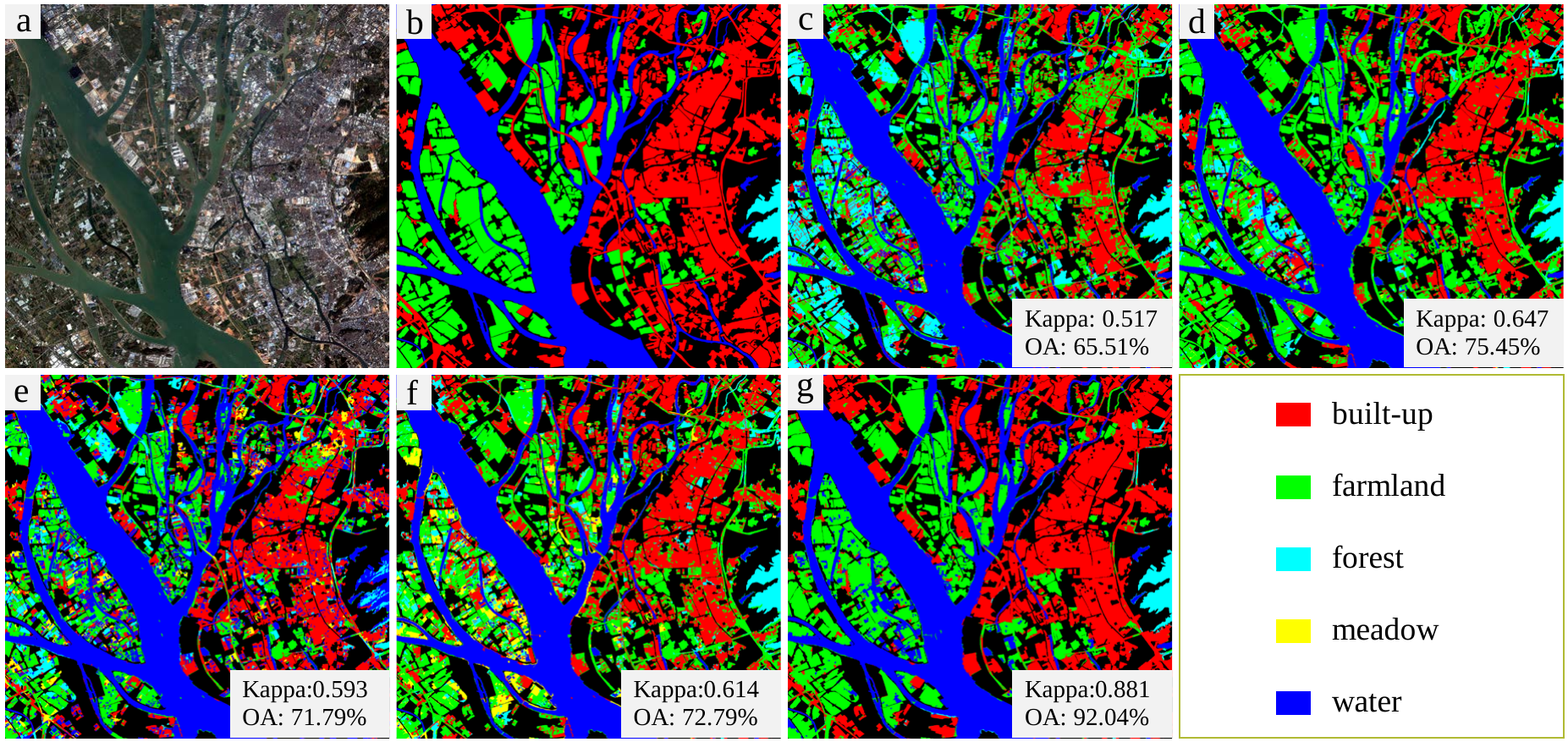}
\caption{Land-cover classification maps of a GF-2 image obtained in Dongguan, Guangdong Province on January 23, 2015. (a) The original image. (b) Ground truth. (c)-(g) Results of MLC+Fusion, RF+Fusion, SVM+Fusion, MLP+Fusion, and PT-GID.}
\label{figure:results-GID-1new}
\end{figure*}

To demonstrate the classification results more intuitively, we display the land-cover classification maps. Fig. \ref{figure:results-GID-1new}(a)-(b) show a GF-2 image belonging to the large-scale classification set, which is obtained in Dongguan, Guangdong Province on January 23, 2015, and its ground truth. Fig. \ref{figure:results-GID-1new} (c)-(g) are the results generated by MLC+Fusion, RF+Fusion, SVM+Fusion, MLP+Fusion, and PT-GID. It can be seen that \emph{farmland} is the most difficult class to be recognized in this image. \emph{Meadow}, \emph{forest} and \emph{farmland} categories are seriously confused by the comparison methods. Compared to the comparison methods, our algorithm generates the best classification performance for \emph{built-up} and \emph{farmland} categories. However, our method misclassified some paddy fields into \emph{water}.

\begin{figure*}[htb!]
\centering
\includegraphics[width=1\textwidth]
{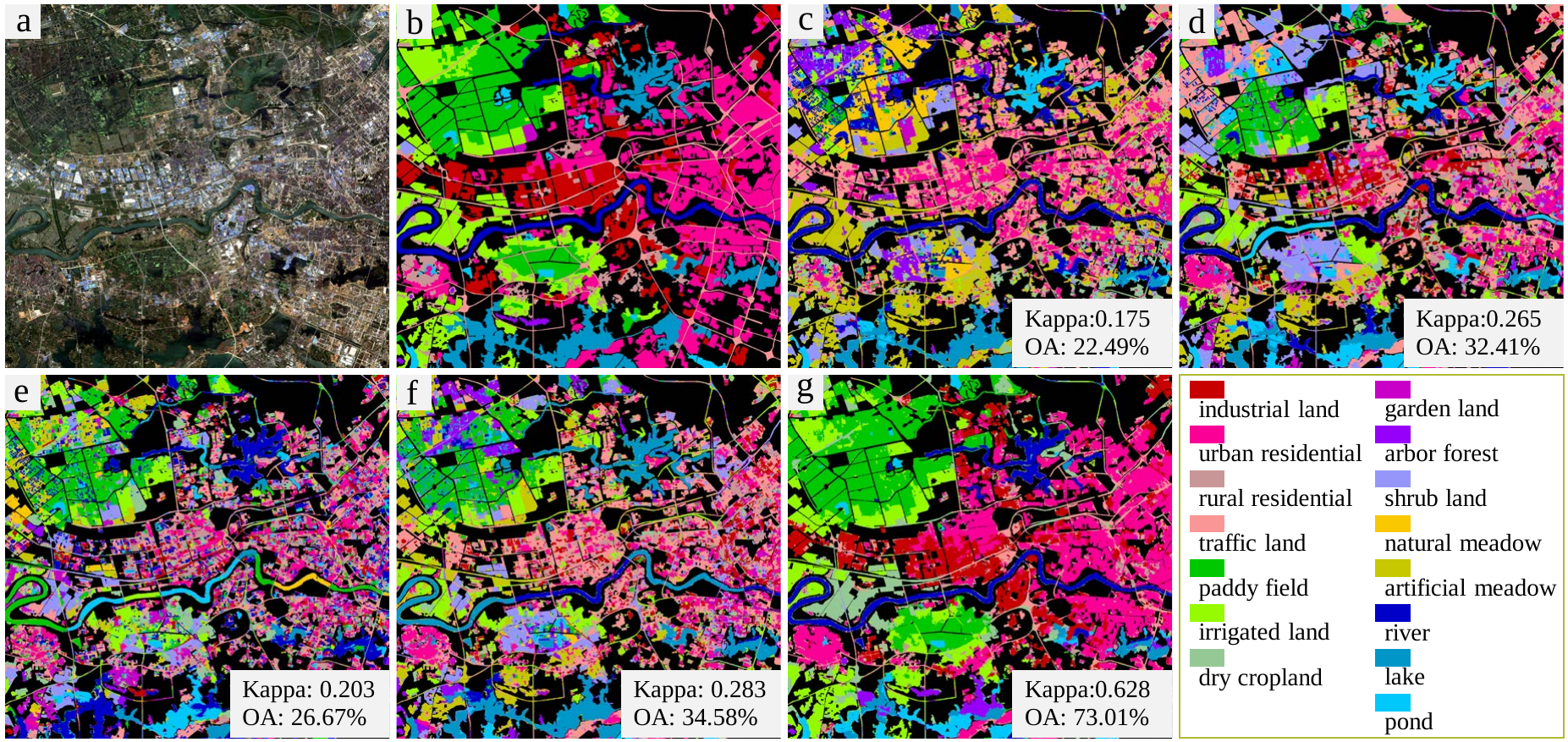}
\caption{Land-cover classification maps of a GF-2 image acquired in Wuhan, Hubei Province on April 11, 2016. (a) The original image. (b) Ground truth. (c)-(g) Results of MLC+Fusion, RF+Fusion, SVM+Fusion, MLP+Fusion, and PT-GID.}
\label{figure:results-GID-2new}
\end{figure*}

Fig. \ref{figure:results-GID-2new}(a) displays a GF-2 image belonging to the fine land-cover classification set, which is acquired in Wuhan, Hubei Province on April 11, 2016, and Fig. \ref{figure:results-GID-2new}(b) is its ground truth. Fig. \ref{figure:results-GID-2new} (c)-(g) show the classification results produced by MLC+Fusion, RF+Fusion, SVM+Fusion, MLP+Fusion, and PT-GID. It can be observed that the comparison methods misclassify large areas of \emph{urban residential} into \emph{rural residential} and \emph{industrial}. This is because the spectral responses of these classes are similar. When the labeling information of the target data is unavailable, it is difficult for the conventional feature extraction methods to represent the contextual properties of these ground objects. Whereas, our scheme generates smooth classification maps that close to the ground truth.

\subsection{Experiments on multi-source images}
\label{sec:multi-source results}

This section focuses on validating the effectiveness of the proposed algorithm on multi-source data. Two deep models are utilized to classify each target image: 1) ResNet-50 pre-trained on the source domain data, 2) ResNet-50 fine-tuned with FT-$\textbf{U}_{tg}$, which are denoted as PT-GID and FT-$\textbf{U}_{tg}$, respectively. For learning transferable models, we set the parameters $\sigma$, $\delta$, $\mu$ to 0.8, 5, 4000 for 5 classes, and to 0.7, 5, 2000 for 15 classes, respectively. Multi-scale information aggregation is utilized, and the initial segmentation size is 400 pixels. We compare our algorithm with object-based classification methods. The fusion of spectral feature, GLCM, DMP, and LBP is used to represent the characteristics of images. The classifiers used are MLC, RF, SVM, and MLP.

\subsubsection{Results of Gaofen-1, Jilin-1, Ziyuan-3, and Sentinel-2A images}

\begin{table*}[b!]
\caption{OA (\%) of different methods on images captured by different sensors.}
\arrayrulecolor{setblue}
\renewcommand\arraystretch{1.1}
\begin{tabular}{cccccccc}
\toprule[1.2pt]
\rowcolor{setgray}
Image&&MLC&RF&SVM&MLP&PT-GID&FT-$\textbf{U}_{tg}$\\
\midrule[0.6pt]
\multirow{2}{*}{GF-1(1)}
&5 classes&62.89&61.95&43.06&68.13&\multicolumn{1}{|c}{82.62}&\textbf{89.84}\\
&15 classes&13.09&21.80&24.33&8.01&\multicolumn{1}{|c}{57.72}&\textbf{60.07}\\
\cmidrule[0.6pt]{1-1}
\multirow{2}{*}{GF-1(2)}
&5 classes&84.87&84.69&55.45&78.99&\multicolumn{1}{|c}{92.40}&\textbf{95.38}\\
&15 classes&14.08&33.46&35.11&25.04&\multicolumn{1}{|c}{74.90}&\textbf{76.89}\\
\cmidrule[0.6pt]{1-1}
\multirow{2}{*}{JL-1(1)}
&5 classes&66.78&67.86&28.26&74.26&\multicolumn{1}{|c}{88.96}&\textbf{90.37}\\
&15 classes&4.49&8.28&24.90&11.57&\multicolumn{1}{|c}{50.12}&\textbf{52.86}\\
\cmidrule[0.6pt]{1-1}
\multirow{2}{*}{JL-1(2)}
&5 classes&80.67&61.14&65.25&67.31&\multicolumn{1}{|c}{72.51}&\textbf{89.08}\\
&15 classes&6.89&23.60&17.47&23.87&\multicolumn{1}{|c}{66.15}&\textbf{71.09}\\
\cmidrule[0.6pt]{1-1}
\multirow{2}{*}{ST-2A(1)}
&5 classes&49.40&76.96&58.10&81.06&\multicolumn{1}{|c}{97.08}&\textbf{97.38}\\
&15 classes&8.30&35.08&8.67&21.11&\multicolumn{1}{|c}{61.14}&\textbf{66.61}\\
\cmidrule[0.6pt]{1-1}
\multirow{2}{*}{ST-2A(2)}
&5 classes&55.34&53.37&45.24&81.58&\multicolumn{1}{|c}{94.46}&\textbf{95.46}\\
&15 classes&9.28&11.22&11.09&2.15&\multicolumn{1}{|c}{28.96}&\textbf{56.89}\\
\cmidrule[0.6pt]{1-1}
\multirow{2}{*}{ZY-3(1)}
&5 classes&63.37&69.24&58.38&62.18&\multicolumn{1}{|c}{85.62}&\textbf{89.21}\\
&15 classes&24.34&34.38&48.18&10.52&\multicolumn{1}{|c}{80.97}&\textbf{82.50}\\
\cmidrule[0.6pt]{1-1}
\multirow{2}{*}{ZY-3(2)}
&5 classes&55.91&72.94&66.74&64.22&\multicolumn{1}{|c}{92.75}&\textbf{94.36}\\
&15 classes&17.27&13.86&30.18&19.32&\multicolumn{1}{|c}{49.47}&\textbf{54.95}\\
\bottomrule[1.2pt]
\end{tabular}
\label{table:onMulti}
\end{table*}

The experimental results of RS images captured by different sensors are shown in Table \ref{table:onMulti}, where OA is used for performance assessment. The accuracy of our algorithm is obviously higher than the comparison methods. For all of the target images, the best OA values of both 5 classes and 15 classes are achieved by FT-$\textbf{U}_{tg}$. These results show that the relevant samples selected from the target domain can strengthen the transferability of CNN models. Therefore, our sample selection and fine-tuning scheme is very effective for multi-source HRRS images.

\begin{figure*}[tb!]
\centering
\includegraphics[width=1\textwidth]
{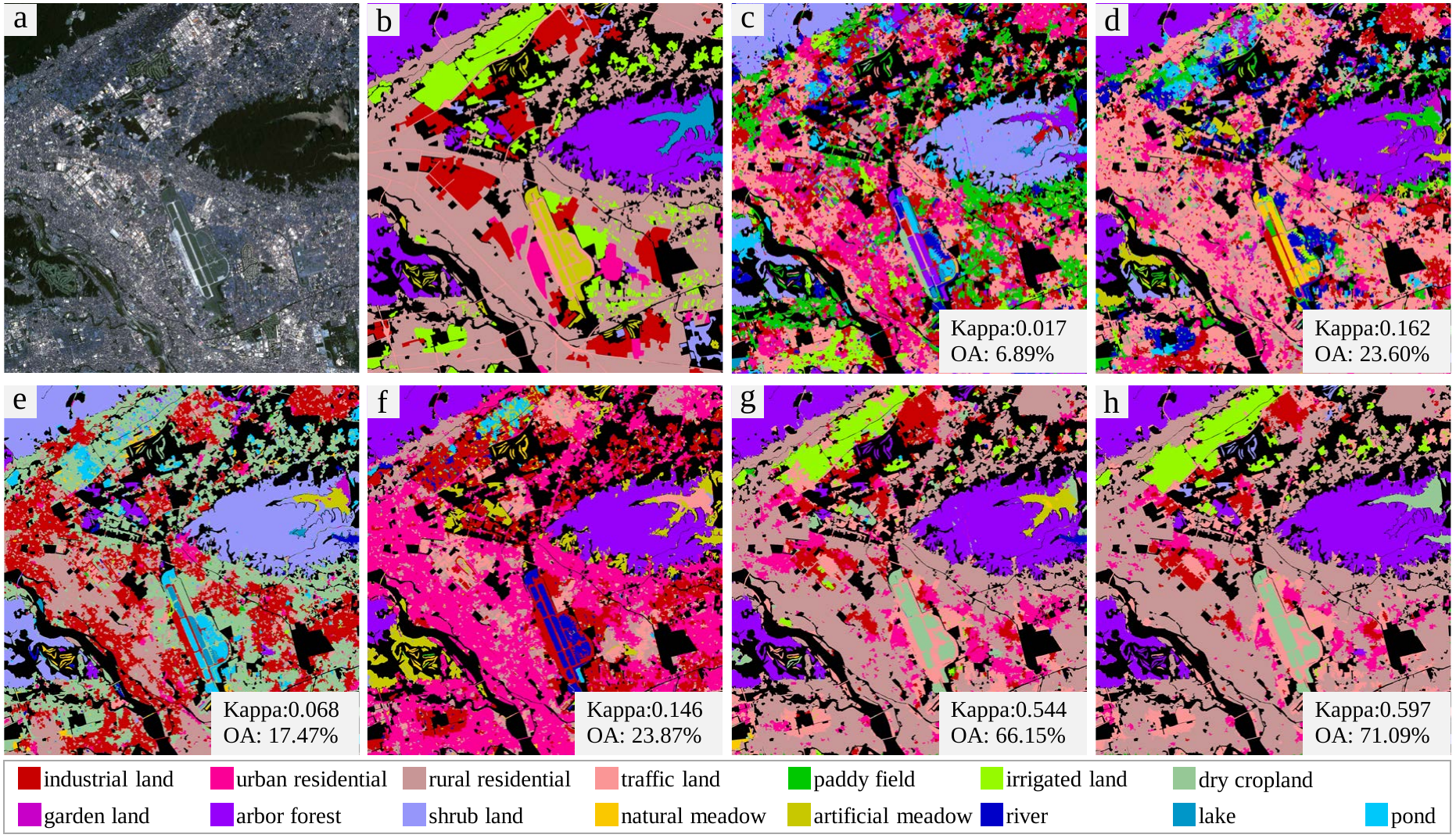}
\caption{Classification results of JL-1(2). (a) The original image. (b) Ground truth. (c)-(h) Results of MLC+Fusion, RF+Fusion, SVM+Fusion, MLP+Fusion, PT-GID, and FT-$\textbf{U}_{tg}$.}
\label{figure:results-Multi-1new}
\end{figure*}

When our algorithm is applied to images acquired by different sensors, FT-$\textbf{U}_{tg}$ can boost the performance compared to PT-GID. Especially for JL-1(2), compared with the results of PT-GID, the OA of FT-$\textbf{U}_{tg}$ increases by 16.57\% on 5 classes, and by 4.94\% on 15 classes. This experimental phenomenon indicates that, if the spectral responses of the target domain and the source domain are similar, the information learned from the source domain samples can benefit the interpretation of the target domain. Conversely, if the spectral responses of the target and source domain are very different ({\em e.g.} obtained by different sensors), the supervision information of the source domain is not reliable for the target domain.

Fig. \ref{figure:results-Multi-1new}(a)-(b) show JL-1(2) and its corresponding ground truth with 15 categories. Fig. \ref{figure:results-Multi-1new} (c)-(h) are the results of MLC+Fusion, RF+Fusion, SVM+Fusion, MLP+Fusion, PT-GID, and FT-$\textbf{U}_{tg}$, respectively. The performance of the comparison methods is unsatisfactory. This is because the distributions of the target domain and the source domain are quite different, and the conventional classifiers do not have sufficient transferability. As shown in Fig. \ref{figure:results-Multi-1new} (g), \emph{urban residential} and \emph{rural residential} are confused, while in Fig. \ref{figure:results-Multi-1new} (h), \emph{urban residential} is correctly classified. The results show that after model fine-tuning, CNN has learnt the distribution of the target domain, which proves that our relevant sample selection scheme can search reliable samples from the target domain.

\begin{figure*}[htb!]
\centering
\includegraphics[width=1\textwidth]
{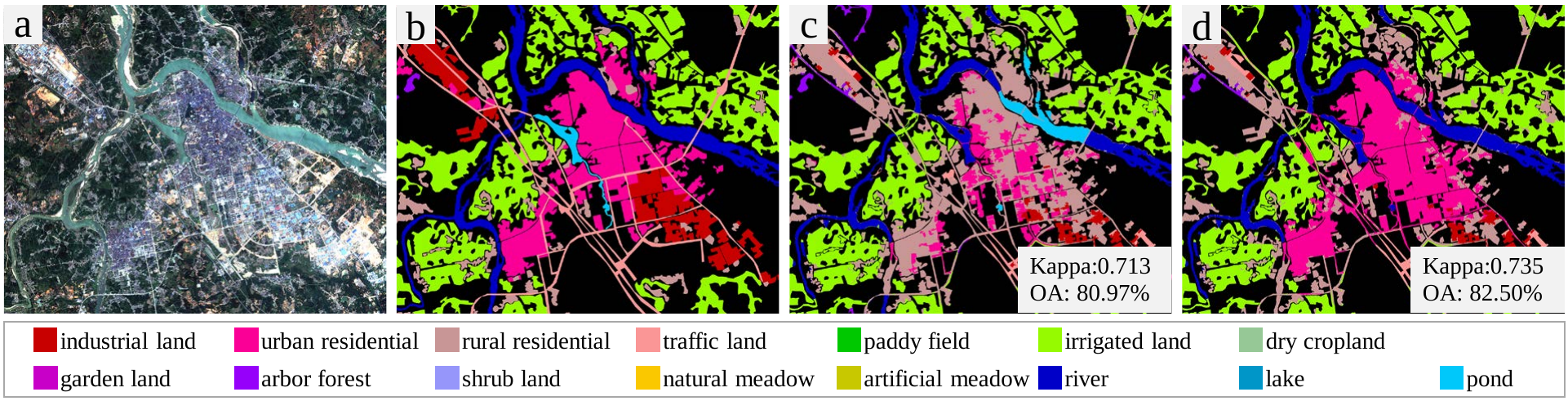}
\caption{Classification results of a sub image in ZY-3(1). (a) The original sub image. (b) Ground truth. (c)-(d) Results of PT-GID and FT-$\textbf{U}_{tg}$.}
\label{figure:results-Multi-2new}
\end{figure*}

Fig. \ref{figure:results-Multi-2new}(a)-(b) display a sub image sized $1000\times 1250$ pixels cropped from ZY-3(1) and its ground truth with 15 categories. The classification results of PT-GID and FT-$\textbf{U}_{tg}$ are demonstrated in Fig. \ref{figure:results-Multi-2new}(c)-(d), respectively. These results show the effect of fine-tuning schemes on the classification performance. Compared to Fig. \ref{figure:results-Multi-2new}(c), less \emph{urban residential} area is mistakenly classified as \emph{rural residential} in Fig. \ref{figure:results-Multi-2new}(d). This is because the spectral responses and the textures of these categories are similar. Whereas, transferred model can learn structural information and spatial relationship of the objects specific to the target domain. These experimental phenomena further validate the robustness and transferability of our approach for diverse HRRS images.

\begin{table*}[b!]
\caption{Comparison of different land-cover classification methods on GE-WH.}
\arrayrulecolor{setblue}
\renewcommand\arraystretch{1.1}
\settowidth\rotheadsize{\theadfont AccuracyQAAAQ}
\begin{tabular}{C{0.5cm}cccccccc}
\toprule[1.2pt]
\rowcolor{setgray}
\multicolumn{2}{>{\columncolor{setgray}}c}{Evaluation Metrics}&&MLC&RF&SVM&MLP&PT-GID&FT-$U_{tg}$\\
\midrule[0.6pt]
\multicolumn{2}{>{}c}{\textbf{Kappa}}&&0.132&0.490&0.253&0.424&\multicolumn{1}{|c}{0.719}&\textbf{0.924}\\
\cmidrule[0.6pt]{1-2}
\multicolumn{2}{>{}c}{\textbf{OA(\%)}}&&27.05&61.43&41.40&58.07&\multicolumn{1}{|c}{80.32}&\textbf{94.56}\\
\cmidrule[0.6pt]{1-2}
\multirow{5}{*}{\rothead{\textbf{\scriptsize \hspace*{1em} User's \hspace*{1em} Accuracy (\%)}}}&Built-up&&58.32&96.05&34.73&80.04&\multicolumn{1}{|c}{\textbf{99.38}}&98.27\\
\cmidrule[0.6pt]{2-2}
&Farmland&&77.51&62.27&61.00&63.49&\multicolumn{1}{|c}{62.65}&\textbf{87.20}\\
\cmidrule[0.6pt]{2-2}
&Forest&&13.89&82.26&32.99&74.90&\multicolumn{1}{|c}{\textbf{97.34}}&97.06\\
\cmidrule[0.6pt]{2-2}
&Meadow&&0.02&0.28&00.30&0.19&\multicolumn{1}{|c}{\textbf{72.69}}&0\\
\cmidrule[0.6pt]{2-2}
&Water&&47.46&74.45&64.46&74.19&\multicolumn{1}{|c}{99.92}&\textbf{99.97}\\
\bottomrule[1.2pt]
\end{tabular}
\label{table:GE}
\end{table*}

\subsubsection{Results of Google Earth platform data in Wuhan}

\begin{figure*}[htb!]
\centering
\includegraphics[width=1\textwidth]
{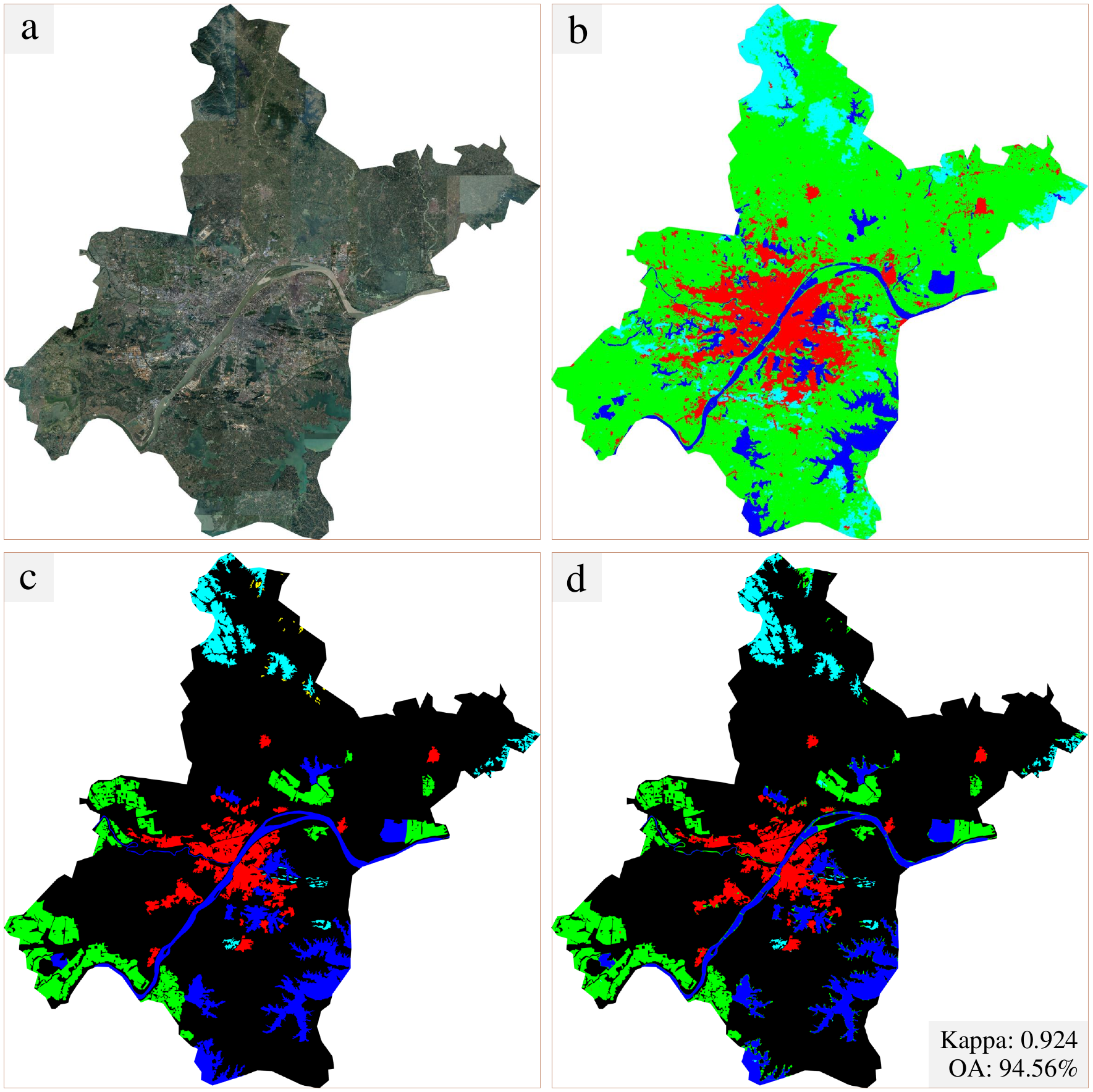}
\caption{(a) GE-WH image. (b) Classification map produced by FT-$\textbf{U}_{tg}$. (c) Partially labeled ground truth. (d) Classification result of FT-$\textbf{U}_{tg}$ in the labeled areas.}
\label{figure:results-GEWH}
\end{figure*}

To test the performance of our method for large-scale land-cover classification, we conduct experiment on GE-WH, which is partially annotated with 5 classes for accuracy assessment, as shown in Fig. \ref{figure:results-GEWH}(c). Table \ref{table:GE} displays the classification performance of the different methods. Overall, our method produces the most satisfactory results. The best result of the comparison methods is generated by RF+Fusion, only reaching Kappa and OA value of 0.490 and 61.43\%, respectively. However, FT-$\textbf{U}_{tg}$ achieves the overall highest Kappa and OA of 0.924 and 94.56\%. Compared to PT-GID, FT-$\textbf{U}_{tg}$ increases Kappa and OA values by 0.205 and 14.24\%, respectively.

Fig. \ref{figure:results-GEWH}(a) shows the original GE-WH. Fig. \ref{figure:results-GEWH}(b) displays the intact classification map produced by FT-$\textbf{U}_{tg}$. Fig. \ref{figure:results-GEWH}(c)-(d) are the partially annotated label mask and the classification result of FT-$\textbf{U}_{tg}$ in the labeled areas, respectively. It can be seen that some areas in Fig. \ref{figure:results-GEWH}(d) are misclassified, for example, \emph{water} area in the middle of Wuhan is identified as \emph{farmland}. However, in the absence of labeling information of the target image, our method achieves Kappa and OA values that exceed 90\%. These results show that our algorithm has the ability to cope with large-coverage HRRS images. Moreover, they also demonstrate the potential of our algorithm to interface with Google Earth and to be applied for practical land-cover mapping.

\section{Sensitivity Analysis}
\label{sec:sensitivity analysis}
In the former section, the experimental results show the promising performance of the proposed method. However, some parameters have impact on the classification results. In this section, we analyse and discuss these factors through additional experiments, including analysis on patch size, segmentation method, and thresholds of transfer learning scheme.

\subsection{Analysis on multi-scale information fusion}
To verify the effectiveness of multi-scale information fusion strategy, we compare the classification performance of the single- and multi-scale methods on the dataset with 5 classes. Image patches of three different sizes are tested: $56\times 56$, $112\times 112$, and $224\times 224$. For selective search segmentation, the initial segmentation size is set to 400. The classification accuracies obtained with different patch sizes are shown in Table \ref{table:I}.

\begin{table*}[htb!]
\caption{Comparison of different patch sample scales.}
\scriptsize
\arrayrulecolor{setblue}
\renewcommand\arraystretch{1.1}
\resizebox{\textwidth}{!}{
\begin{tabular}{cccccccccc}
\toprule[1.2pt]
\rowcolor{setgray}
\textbf{Patch}&&\textbf{Kappa}&\textbf{OA}&&\multicolumn{5}{>{\columncolor{setgray}}c}{\textbf{User's Accuracy (\%)}}\\
\cmidrule[0.6pt]{6-10}
\rowcolor{setgray}
\textbf{Size}&&&\textbf{(\%)}&&built-up&farmland&forest&meadow&water\\
\midrule[0.6pt]
$56\times56$&&0.915&95.41&&80.02&89.47&76.83&\textbf{78.87}&89.12\\
\cmidrule[0.6pt]{1-1}
$112\times112$&&0.892&94.36&&71.33&87.10&76.31&77.84&\textbf{89.71}\\
\cmidrule[0.6pt]{1-1}
$224\times224$&&0.880&92.67&&72.13&86.41&75.95&71.54&87.75\\
\cmidrule[0.6pt]{1-1}
Multi-scale&&\textbf{0.924}&\textbf{96.28}&&\textbf{88.42}&\textbf{91.85}&\textbf{79.42}&70.55&87.60\\
\bottomrule[1.2pt]
\end{tabular}}
\label{table:I}
\end{table*}

For single scales, the optimal result is achieved by the smallest patch size $56\times 56$. This is because our classification method is based on the image patches generated from non-overlapping grid partition, and all pixels in a patch are assigned with the same label. If the patch size is too large, the object details in the patches will be lost. Compared with the best results given by the single-scale approaches, multi-scale information fusion strategy attains the highest Kappa and OA of 0.924 and 96.28\%, respectively. These results indicate that ground objects in HRRS images show great variations of contextual information in different scales. And combining image information of different scales helps to characterize the spatial distributions of the ground objects.

\subsection{Analysis on segmentation}

\begin{figure*}[htb!]
\centering
\includegraphics[width=0.65\textwidth]
{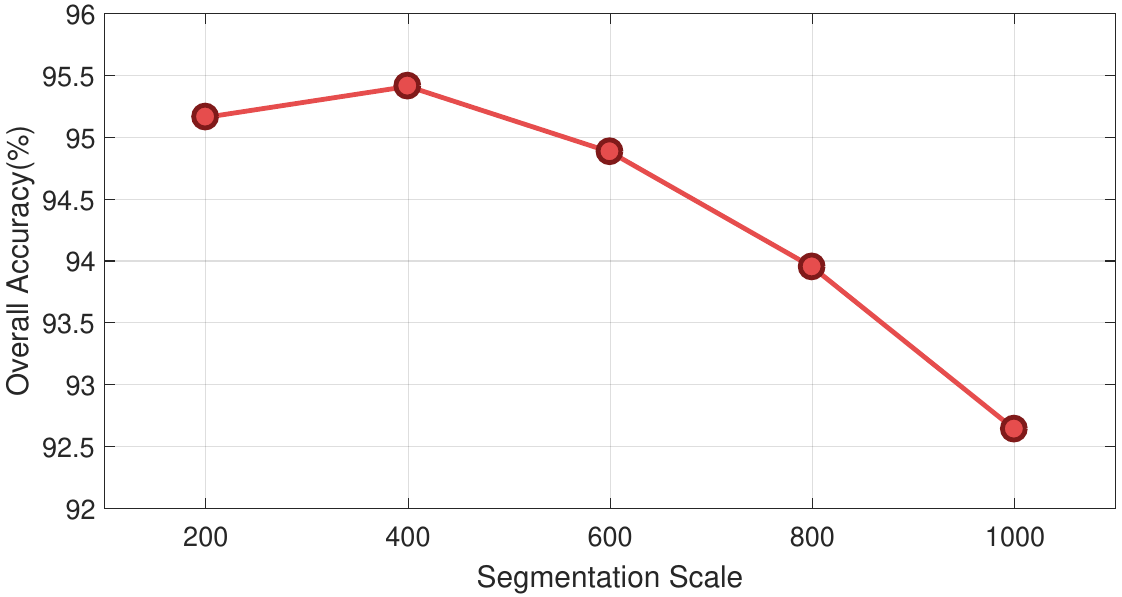}
\caption{Sensitivity analysis for the segmentation scale.}
\label{figure:segmentation}
\end{figure*}

To analyse the influence of the segmentation scale, we test five different initial segmentation sizes of selective search method, including 200, 400, 600, 800, and 1000, on the validation set with 5 classes. The image patches of size $56\times 56$ are used for model pre-training and patch-wise classification. The mean OA values generated by each segmentation scale are illustrated in Fig. \ref{figure:segmentation}.

It can be seen that, the mean OA value first slightly increases and then continuously decreases with the increase of the segmentation scale. When the segmentation scale is set to 400, the best result is yielded. In general, for object-based land-cover classification, the most suitable segmentation scale depends on the spatial resolution of the RS image and the characteristics of the ground objects. If the image is over-segmented, more noise will be introduced into the classification results, and if the image is under-segmented, the classification map will lose a lot of details.

In addition, we compare the performance of selective search method with multi-resolution segmentation on 5 classes. We utilize the multi-resolution segmentation function embedded in eCognition, which is a professional and efficient software for RS image analysis, to generate segmentation maps. The scale parameter is set to 400, the same with that of selective search method. And the patch size is set to $56\times 56$. The results are shown in Table \ref{table:segMethod}.

Selective search provides slightly higher Kappa and OA value than the results of multi-resolution segmentation. On \emph{farmland} and \emph{water}, selective search behaves better, while multi-resolution segmentation generates higher accuracy on \emph{built-up}, \emph{forest} and \emph{meadow} categories. The experimental results show that the performance of the two methods is comparable. Since this procedure is not the key factor to determine the accuracy of our approach, one may employ either selective search or multi-resolution segmentation method.

\begin{table*}[htb!]
\caption{Comparison of selective search and multi-resolution segmentation.}
\arrayrulecolor{setblue}
\renewcommand\arraystretch{1.1}
\resizebox{\textwidth}{!}{
\begin{tabular}{cccccccccc}
\toprule[1.2pt]
\rowcolor{setgray}
\textbf{Segmentation}&&\textbf{Kappa}&\textbf{OA}&
&\multicolumn{5}{>{\columncolor{setgray}}c}{\textbf{User's Accuracy (\%)}}\\
\cmidrule[0.6pt]{6-10}
\rowcolor{setgray}
\textbf{Method}&&&\textbf{(\%)}&&built-up&farmland&forest&meadow&water\\
\midrule[0.6pt]
Selective Search Method&&\textbf{0.915}&\textbf{95.41}&&80.02&\textbf{89.47}&76.83&78.87&\textbf{89.12}\\
\cmidrule[0.6pt]{1-1}
Multi-resolution Segmentation&&0.907&94.62&&\textbf{82.35}&87.50&\textbf{81.89}&\textbf{83.94}&88.01\\
\bottomrule[1.2pt]
\end{tabular}}
\label{table:segMethod}
\end{table*}

\subsection{Analysis on transfer learning parameters}
\begin{figure*}[htb!]
\subfigure[$\sigma$ (5 classes)]
{\includegraphics[width=0.32\textwidth]
{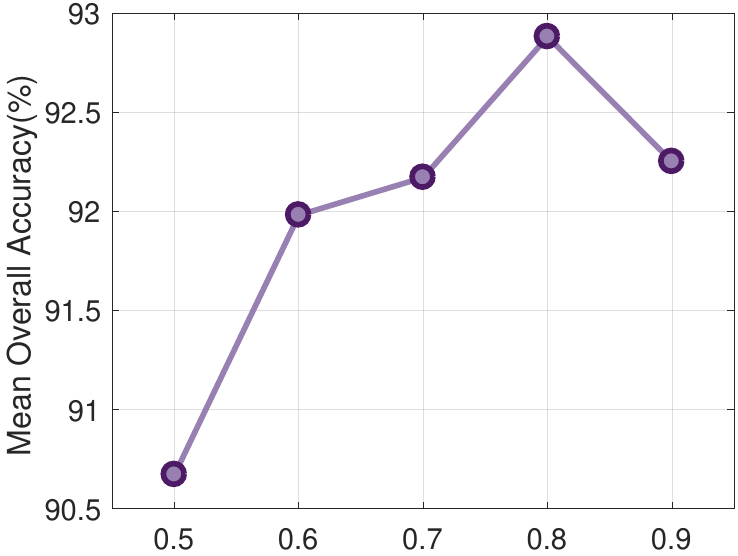}}
\subfigure[$\delta$ (5 classes)]
{\includegraphics[width=0.32\textwidth]
{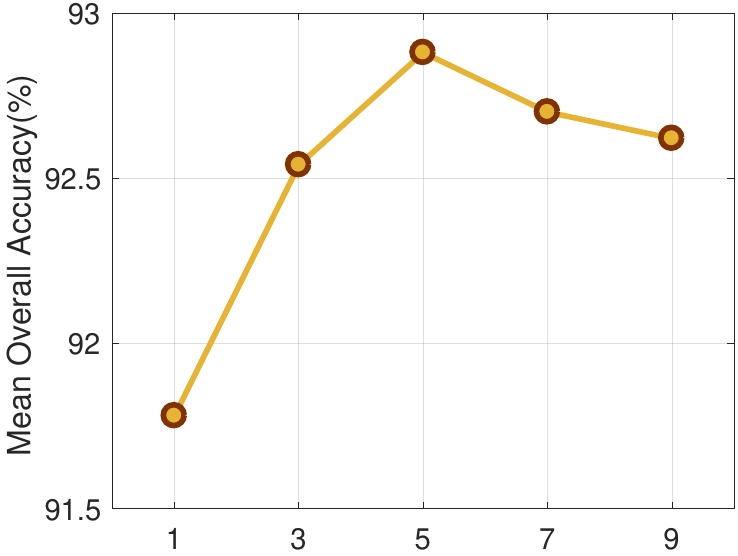}}
\subfigure[$\mu$ (5 classes)]
{\includegraphics[width=0.32\textwidth]
{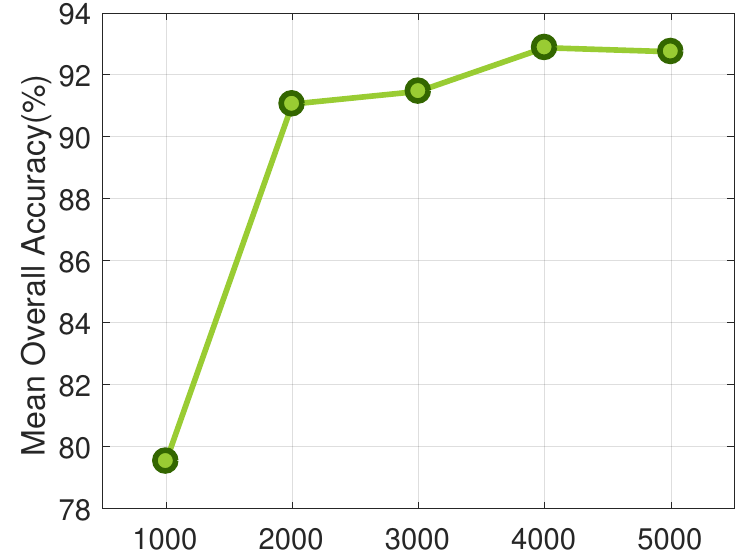}}\\
\subfigure[$\sigma$ (15 classes)]
{\includegraphics[width=0.32\textwidth]
{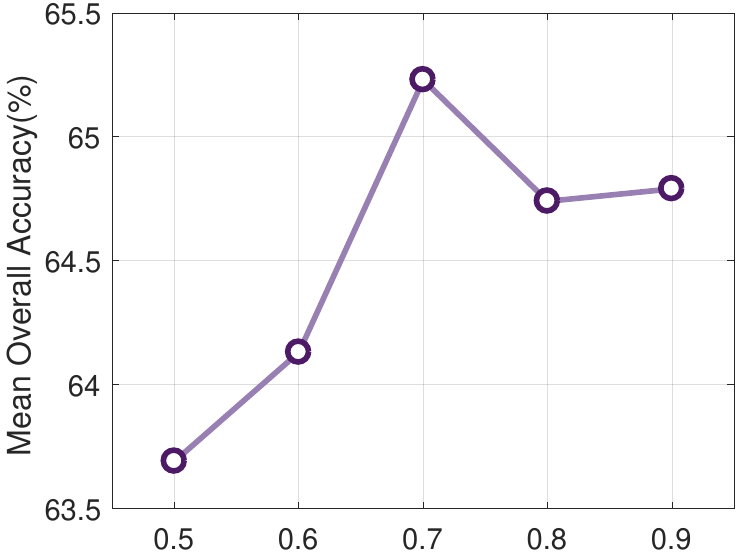}}
\subfigure[$\delta$ (15 classes)]
{\includegraphics[width=0.32\textwidth]
{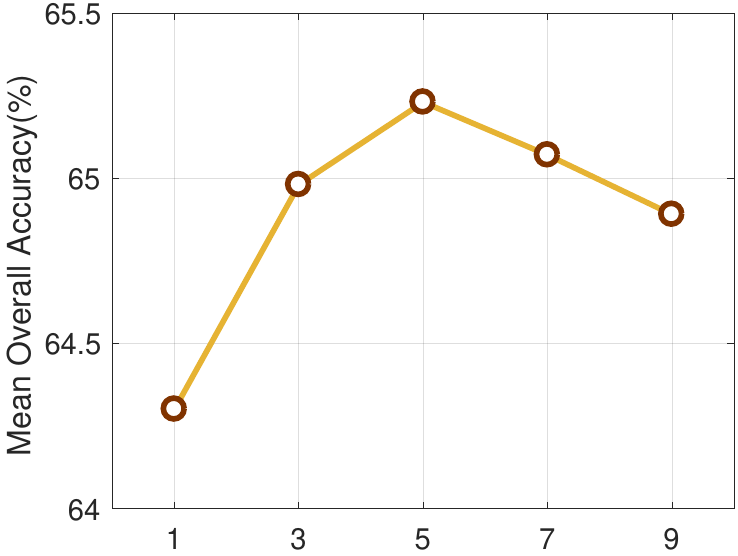}}
\subfigure[$\mu$ (15 classes)]
{\includegraphics[width=0.32\textwidth]
{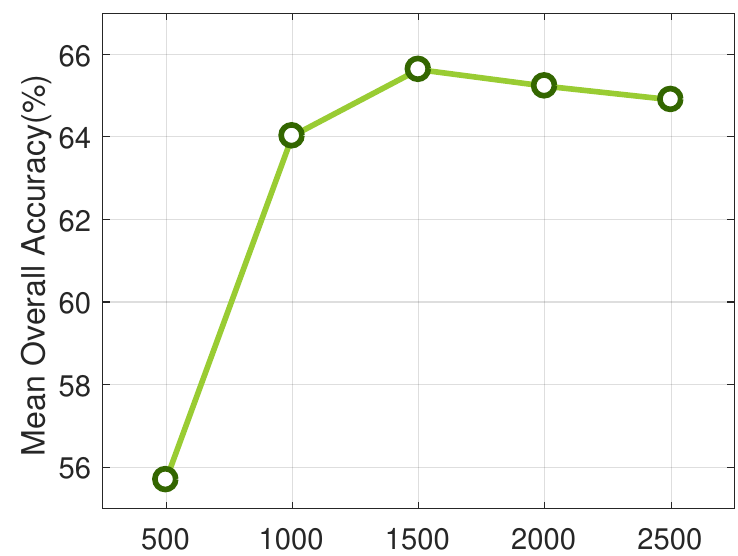}}
\caption{Sensitivity analysis for $\sigma$, $\delta$, and $\mu$ parameters.}
\label{figure:SA2}
\end{figure*}

To investigate how the $\sigma$, $\delta$, and $\mu$ parameters affect the proposed transfer learning scheme, we study on every parameter separately, i.e. at each time experimenting with one parameter and fixing the other two. Tests are conducted on GF-1, JL-1, ZY-3 and ST-2A images, and the mean OA value of these images are calculated to indicate the impact of the parameters. The relationship between the mean OA value and $\sigma$, $\delta$, $\mu$ is presented in Fig. \ref{figure:SA2}.

Parameter $\sigma$ is varied from 0.5 to 0.9, with an interval of 0.1. $\delta$ and $\mu$ are set to 5 and 4000 for 5 classes, 5 and 2000 for 15 classes, respectively. $\sigma$ is the threshold for pseudo-label assignment. In the transfer learning scheme, candidate patches with classification probability greater than $\sigma$ are assigned the pseudo-labels. When $\sigma$ is set to 0.8 and 0.7, the best performance is achieved on 5 classes and 15 classes respectively. This is because smaller $\sigma$ value leads to unreliable pseudo-labels, while larger $\sigma$ value may result in very few candidate samples, which are insufficient to train a well-performed deep model. Furthermore, when the classes are finer, the same category of different data sources have a larger difference in characteristics, and higher confidence may filter out valuable samples.

Parameter $\delta$ is varied from 1 to 9, with an interval of 2. $\sigma$ and $\mu$ are set to 0.8 and 4000 for 5 classes, 0.7 and 2000 for 15 classes, respectively. $\delta$ is utilized to filter out the pseudo-labels that do not match the annotation information of the source domain. If the true-labels of the top $\delta$ retrieved samples are identical to the pseudo-label of the query patch, this query patch is retained, otherwise it is removed. The $\delta$ with value of 5 provides the highest mean OA value. The reason for this phenomenon is that smaller $\delta$ value can not guarantee the consistency of the pseudo-labels and the true-labels, while larger $\delta$ is too strict to extract sufficient candidate patches from the target domain.

Parameter $\mu$ is varied from 1000 to 5000, with an interval of 1000 on 5 classes, and from 500 to 2500, with an interval of 500 on 15 classes. $\sigma$ and $\delta$ are set to 0.8 and 5 for 5 classes, 0.7 and 5 for 15 classes, respectively. $\mu$ is employed to limit the number of target domain samples. It can be observed that, on 5 classes, at the beginning, the classification accuracy increases with the increase of the $\mu$ value, and when $\mu$ is greater than 2000, the mean OA value rises more gently. The highest accuracy is achieved when $\mu$ is equal to 4000. This shows that $\mu$ of 2000 is enough to select sufficient samples for training a well-performed model, although it does not provide the best performance. And when $\mu$ is larger than 4000, there is redundancy between the selected target domain samples, which causes the deep model to be biased toward the redundant information. In addition, excessive training samples distinctly reduce the model training efficiency. On 15 categories, our approach behaves best when $\mu$ is equal to 1500. This is because the 5 major categories are subdivided in the fine land-cover classification set, and each sub-class contains fewer samples.

The high diversity of the samples in the source domain (i.e. GID) enables the pre-trained CNN to be discriminating. Therefore, diverse target samples can be correctly identified and selected by the models. To ensure that sufficient samples are extracted for model fine-tuning, we have not adopted diverse criterion to deal with data redundancy. Due to the large volume of CNN's parameters and the variety of target samples, appropriate data redundancy does not have a significant impact on model performance. In the future research, we are interested in further investigating how to address the redundancy problem with diverse criterions.

\section{Discussion}
\label{sec:discussion}
Land-cover classification is closely tied to the ecological condition of the Earth's surface and have significant implications for global ecosystem health, water quality, and sustainable land management. Most studies on large-scale land-cover classification generally use the low-/medium-spatial resolution RS images, however, due to the lack of spatial information, these images are insufficient for detailed mapping for high heterogeneous areas \cite{limitation}. By contrast, high-spatial resolution images provide rich texture, shape, and spatial distribution information of ground objects, which contribute significantly to distinguish categories with similar spectral characteristics. Nevertheless, because of the narrow spatial coverage and high economic costs, high-spatial resolution images are commonly employed in land-cover classification for some specific small regions. In addition, even if a mass of HRRS images are available, in the case where accurate annotation is difficult to quickly obtain, the classifiers will have insufficient adaptability to data and connot be used in practical applications. Therefore, it is highly demanded to develop robust and transferable algorithms to achieve high-precision land-cover classification at large-scale.

Considering that the structure and spatial relationship of the ground objects do not change with the acquisition conditions, we relate the intrinsic characteristics of the objects in the multi-source data through the high-level deep features of the images. Our approach is inspired by pseudo-label assignment \cite{PseudoLabel1,PseudoLabel2} and joint fine-tuning \cite{JointFinetune1,JointFinetune2} methods. However, compared with these two methods, our semi-supervised algorithm does not need any annotation information of the target domain, and the reliability of our method is improved by the constraint of feature similarity. Our approach achieves complete automatic classification for the unlabeled target images, providing new possibilities for real-time land-cover classification applications. Although our approach proves to be effective in experiments and presents remarkable performance on 5 classes, for more complex categories, the classification accuracy still has room for improvement (see Table \ref{table:onGID}). We use confusion matrices to analyse the behavior of our method on different fine categories.

\begin{figure*}[htb!]
\subfigure[\scriptsize PT-GID (GID)]{
\includegraphics[height=0.24\textheight]
{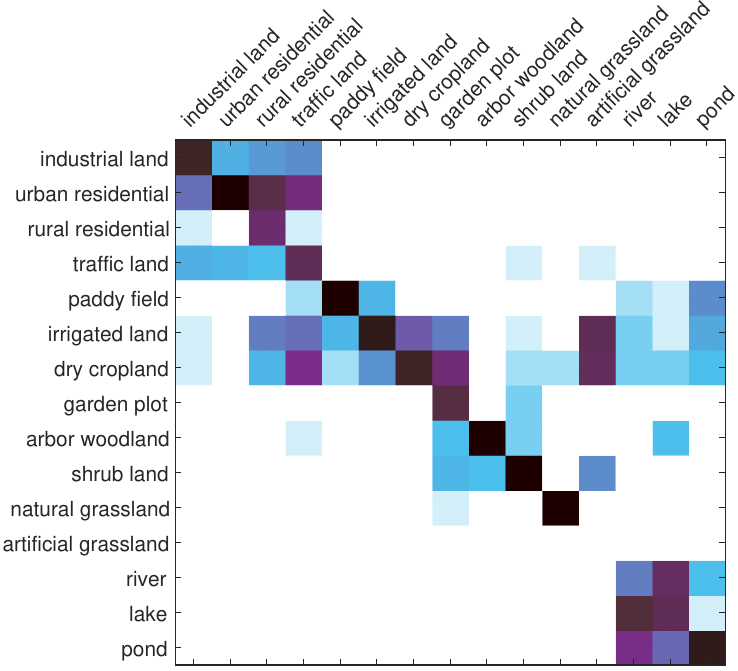}}\hskip -3pt
\subfigure[\scriptsize PT-GID (multi-source)]{
\includegraphics[height=0.24\textheight]
{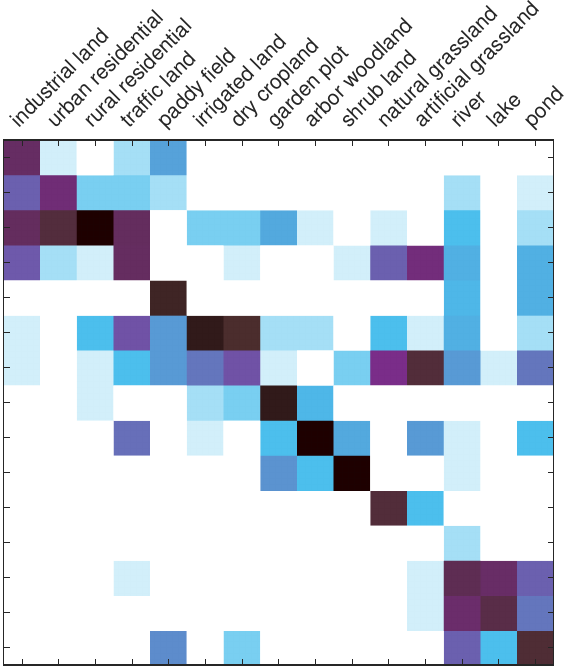}}\hskip -3pt
\subfigure[\scriptsize FT-$\textbf{U}_{tg}$ (multi-source)]{
\includegraphics[height=0.24\textheight]
{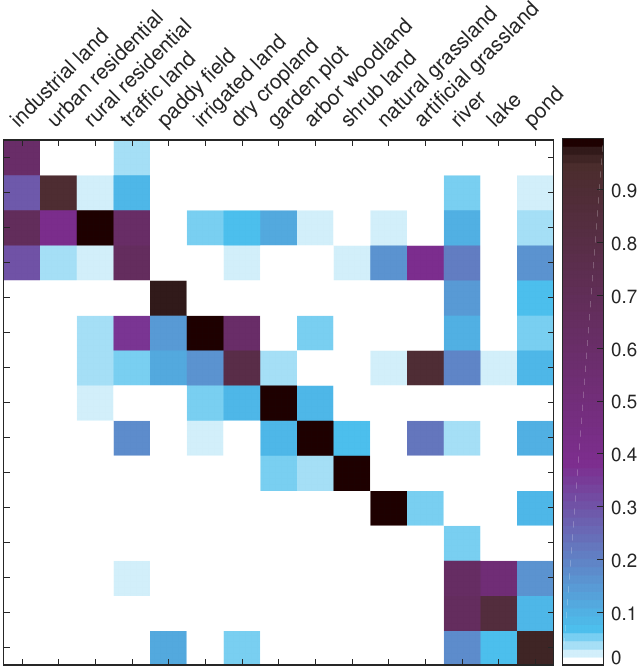}}
\caption{Confusion matrices of land-cover classification results on GID and multi-source images with 15 categories.}
\label{figure:ConfusionMatrix}
\end{figure*}

As shown in Fig. \ref{figure:ConfusionMatrix}, (a)-(c) are the confusion matrices of classification results of PT-GID on GID, PT-GID on multi-source images, and FT-$\textbf{U}_{tg}$  on multi-source images, respectively. It can be seen that the fine categories of the same major class are seriously confused, for instance, \emph{river}, \emph{lake}, and \emph{pond} are misclassified. In addition, \emph{farmland} and \emph{meadow} are severely confused. This is because that they are difficult to be recognized based on spectral, texture and structure information. Furthermore, the shapes of \emph{water} and \emph{farmland} areas are diverse, and they present different appearances in different seasons. In this case, multi-temporal analysis may provide the information that distinguishes between these categories \cite{multi-temporal1,multi-temporal2,multi-temporal3,multi-temporal4}. However, multi-temporal analysis generally requires explicit labels of the samples as supervised information for multi-temporal feature learning. Currently, our method cannot guarantee that the pseudo-labels of the selected samples are sufficiently accurate. Therefore, in our future work, it is of great interest to us to further study multi-supervised multi-temporal analysis.

\section{Conclusion}
\label{sec:conclusion}
We present a land-cover classification algorithm that can be applied to classify multi-source HRRS images. The proposed algorithm has the following attractive properties: 1) it automatically selects training samples from the target domain based on the contextual information extracted from deep model. In consequence, it does not require new manual annotation or algorithm adjustment when being applied to multi-source images. 2) it uses multi-scale contextual information for classification. Therefore, the spatial distributions of the objects are characterized, and the transferability of deep models for RS images with different resolutions is strengthen. 3) it combines patch-wise classification and hierarchical segmentation. The accurate category and boundary information is simultaneously obtained, and the classification noise is reduced in the classification map.

In addition, we constructed a large-scale land-cover dataset, {\em i.e.} GID, with 150 high-resolution GF-2 images. It well represents the true distribution of land-cover categories and can be used to train CNN model specific to RS data. We conduct experiments on multi-source HRRS images, including Gaofen-2 (GF-2) images in GID, coupled with Gaofen-1 (GF-1), Jilin-1 (JL-1), Ziyuan-3 (ZY-3), Sentinel-2A (ST-2A), and Google Earth (GE-WH) platform data. The proposed algorithm shows encouraging classification performance. To benefit researchers, GID have been provided online at \url{https://x-ytong.github.io/project/GID.html}.

\bibliographystyle{IEEEtran}
\bibliography{reference}

\begin{thebibliography}{100}
\providecommand{\url}[1]{#1}
\csname url@samestyle\endcsname
\providecommand{\newblock}{\relax}
\providecommand{\bibinfo}[2]{#2}
\providecommand{\BIBentrySTDinterwordspacing}{\spaceskip=0pt\relax}
\providecommand{\BIBentryALTinterwordstretchfactor}{4}
\providecommand{\BIBentryALTinterwordspacing}{\spaceskip=\fontdimen2\font plus
\BIBentryALTinterwordstretchfactor\fontdimen3\font minus
  \fontdimen4\font\relax}
\providecommand{\BIBforeignlanguage}[2]{{%
\expandafter\ifx\csname l@#1\endcsname\relax
\typeout{** WARNING: IEEEtran.bst: No hyphenation pattern has been}%
\typeout{** loaded for the language `#1'. Using the pattern for}%
\typeout{** the default language instead.}%
\else
\language=\csname l@#1\endcsname
\fi
#2}}
\providecommand{\BIBdecl}{\relax}
\BIBdecl

\bibitem{application1}
R.~Mathieu, C.~Freeman, and J.~Aryal, ``Mapping private gardens in urban areas
  using object-oriented techniques and very high-resolution satellite
  imagery,'' \emph{Landscape and Urban Planning}, vol.~81, no.~3, pp. 179--192,
  2007.

\bibitem{application2}
H.~Shi, L.~Chen, F.-k. Bi, H.~Chen, and Y.~Yu, ``Accurate urban area detection
  in remote sensing images,'' \emph{{IEEE} Geoscience and Remote Sensing
  Letters}, vol.~12, no.~9, pp. 1948--1952, 2015.

\bibitem{application3}
A.~Ozdarici-Ok, A.~O. Ok, and K.~Schindler, ``Mapping of agricultural crops
  from single high-resolution multispectral images¡ªdata-driven smoothing vs.
  parcel-based smoothing,'' \emph{Remote Sensing}, vol.~7, no.~5, pp.
  5611--5638, 2015.

\bibitem{application4}
C.~Zhang and J.~M. Kovacs, ``The application of small unmanned aerial systems
  for precision agriculture: a review,'' \emph{Precision Agriculture}, vol.~13,
  no.~6, pp. 693--712, 2012.

\bibitem{application5}
J.~P. Ardila, V.~A. Tolpekin, W.~Bijker, and A.~Stein, ``Markov random
  field-based super-resolution mapping for identification of urban trees in vhr
  images,'' \emph{ISPRS J. Photogrammetry and Remote Sensing}, vol.~66, no.~6,
  pp. 762--775, 2011.

\bibitem{application6}
M.~Fauvel, Y.~Tarabalka, J.~A. Benediktsson, J.~Chanussot, and J.~C. Tilton,
  ``Advances in spectral-spatial classification of hyperspectral images,''
  \emph{Proceedings of the {IEEE}}, vol. 101, no.~3, pp. 652--675, 2013.

\bibitem{source}
G.~Moser, S.~B. Serpico, and J.~A. Benediktsson, ``Land-cover mapping by markov
  modeling of spatial--contextual information in very-high-resolution remote
  sensing images,'' \emph{Proceedings of the {IEEE}}, vol. 101, no.~3, pp.
  631--651, 2013.

\bibitem{separability}
L.~Bruzzone and L.~Carlin, ``A multilevel context-based system for
  classification of very high spatial resolution images,'' \emph{{IEEE} Trans.
  on Geoscience and Remote Sensing}, vol.~44, no.~9, pp. 2587--2600, 2006.

\bibitem{DAreview}
D.~Tuia, C.~Persello, and L.~Bruzzone, ``Domain adaptation for the
  classification of remote sensing data: An overview of recent advances,''
  \emph{{IEEE} Geoscience and Remote Sensing Magazine}, vol.~4, no.~2, pp.
  41--57, 2016.

\bibitem{spectral1}
J.~R. Jensen and K.~Lulla, ``Introductory digital image processing: A remote
  sensing perspective,'' \emph{Geocarto International}, vol.~2, no.~1, pp.
  65--65, 1986.

\bibitem{spectral2}
P.~Gong, D.~J. Marceau, and P.~J. Howarth, ``A comparison of spatial feature
  extraction algorithms for land-use classification with spot hrv data,''
  \emph{Remote Sensing of Environment}, vol.~40, no.~2, pp. 137--151, 1992.

\bibitem{spectral3}
P.~Casals-Carrasco, S.~Kubo, and B.~B. Madhavan, ``Application of spectral
  mixture analysis for terrain evaluation studies,'' \emph{International
  Journal of Remote Sensing}, vol.~21, no.~16, pp. 3039--3055, 2000.

\bibitem{spatialSpectral1}
S.~Giada, T.~De~Groeve, D.~Ehrlich, and P.~Soille, ``Information extraction
  from very high resolution satellite imagery over lukole refugee camp,
  tanzania,'' \emph{International Journal of Remote Sensing}, vol.~24, no.~22,
  pp. 4251--4266, 2003.

\bibitem{spatialSpectral2}
Y.~Tarabalka, J.~Chanussot, and J.~A. Benediktsson, ``Segmentation and
  classification of hyperspectral images using minimum spanning forest grown
  from automatically selected markers,'' \emph{{IEEE} Trans. on Systems, Man,
  and Cybernetics, Part B (Cybernetics)}, vol.~40, no.~5, pp. 1267--1279, 2010.

\bibitem{spatialSpectral3}
Y.~Tarabalka, M.~Fauvel, J.~Chanussot, and J.~A. Benediktsson, ``Svm-and
  mrf-based method for accurate classification of hyperspectral images,''
  \emph{{IEEE} Geoscience and Remote Sensing Letters}, vol.~7, no.~4, pp.
  736--740, 2010.

\bibitem{spatialSpectral4}
Y.~Zhong, J.~Zhao, and L.~Zhang, ``A hybrid object-oriented conditional random
  field classification framework for high spatial resolution remote sensing
  imagery,'' \emph{{IEEE} Trans. on Geoscience and Remote Sensing}, vol.~52,
  no.~11, pp. 7023--7037, 2014.

\bibitem{spatialSpectral5}
L.~Ma, M.~Li, X.~Ma, L.~Cheng, P.~Du, and Y.~Liu, ``A review of supervised
  object-based land-cover image classification,'' \emph{ISPRS J. Photogrammetry
  and Remote Sensing}, vol. 130, pp. 277--293, 2017.

\bibitem{semanticGap1}
B.~Zhao, Y.~Zhong, G.-S. Xia, and L.~Zhang, ``Dirichlet-derived multiple topic
  scene classification model for high spatial resolution remote sensing
  imagery,'' \emph{{IEEE} Trans. on Geoscience and Remote Sensing}, vol.~54,
  no.~4, pp. 2108--2123, 2016.

\bibitem{semanticGap2}
Y.~Zhong, S.~Wu, and B.~Zhao, ``Scene semantic understanding based on the
  spatial context relations of multiple objects,'' \emph{Remote Sensing},
  vol.~9, no.~10, p. 1030, 2017.

\bibitem{hu2016fast}
F.~Hu, G.-S. Xia, J.~Hu, Y.~Zhong, and K.~Xu, ``Fast binary coding for the
  scene classification of high-resolution remote sensing imagery,''
  \emph{Remote Sensing}, vol.~8, no.~7, p. 555, 2016.

\bibitem{yu2016color}
H.~Yu, W.~Yang, G.-S. Xia, and G.~Liu, ``A color-texture-structure descriptor
  for high-resolution satellite image classification,'' \emph{Remote Sensing},
  vol.~8, no.~3, p. 259, 2016.

\bibitem{shao2013extreme}
W.~Shao, W.~Yang, and G.-S. Xia, ``Extreme value theory-based calibration for
  the fusion of multiple features in high-resolution satellite scene
  classification,'' \emph{International Journal of Remote Sensing}, vol.~34,
  no.~23, pp. 8588--8602, 2013.

\bibitem{hu2017deep}
F.~Hu, G.-S. Xia, and L.~Zhang, ``Deep sparse representations for land-use
  scene classification in remote sensing images,'' in \emph{IEEE International
  Conference on Signal Processing}, 2017, pp. 192--197.

\bibitem{yang2015learning}
W.~Yang, X.~Yin, and G.-S. Xia, ``Learning high-level features for satellite
  image classification with limited labeled samples,'' \emph{{IEEE} Trans. on
  Geoscience and Remote Sensing}, vol.~53, no.~8, pp. 4472--4482, 2015.

\bibitem{SceneClassification1}
F.~Hu, G.-S. Xia, J.~Hu, and L.~Zhang, ``Transferring deep convolutional neural
  networks for the scene classification of high-resolution remote sensing
  imagery,'' \emph{Remote Sensing}, vol.~7, no.~11, pp. 14\,680--14\,707, 2015.

\bibitem{deepReview}
X.~X. Zhu, D.~Tuia, L.~Mou, G.-S. Xia, L.~Zhang, F.~Xu, and F.~Fraundorfer,
  ``Deep learning in remote sensing: A comprehensive review and list of
  resources,'' \emph{{IEEE} Geoscience and Remote Sensing Magazine}, vol.~5,
  no.~4, pp. 8--36, 2017.

\bibitem{AlexNet}
A.~Krizhevsky, I.~Sutskever, and G.~E. Hinton, ``Imagenet classification with
  deep convolutional neural networks,'' in \emph{International Conference on
  Neural Information Processing Systems}, 2012, pp. 1097--1105.

\bibitem{visualizing}
M.~D. Zeiler and R.~Fergus, ``Visualizing and understanding convolutional
  networks,'' in \emph{European Conference on Computer Vision}.\hskip 1em plus
  0.5em minus 0.4em\relax Springer, 2014, pp. 818--833.

\bibitem{SceneClassification2}
G.-S. Xia, J.~Hu, F.~Hu, B.~Shi, X.~Bai, Y.~Zhong, L.~Zhang, and X.~Lu, ``Aid:
  A benchmark data set for performance evaluation of aerial scene
  classification,'' \emph{{IEEE} Trans. on Geoscience and Remote Sensing},
  vol.~55, no.~7, pp. 3965--3981, 2017.

\bibitem{ObjectDetection}
G.-S. Xia, X.~Bai, J.~Ding, Z.~Zhu, S.~Belongie, J.~Luo, M.~Datcu, M.~Pelillo,
  and L.~Zhang, ``Dota: A large-scale dataset for object detection in aerial
  images,'' in \emph{IEEE Conference on Computer Vision and Pattern
  Recognition}, 2018.

\bibitem{Retrieval}
P.~Napoletano, ``Visual descriptors for content-based retrieval of
  remote-sensing images,'' \emph{International Journal of Remote Sensing},
  vol.~39, no.~5, pp. 1343--1376, 2018.

\bibitem{Jiang2017}
T.-B. Jiang, G.-S. Xia, Q.-K. Lu, and W.-M. Shen, ``Retrieving aerial scene
  images with learned deep image-sketch features,'' \emph{Journal of Computer
  Science and Technology}, vol.~32, no.~4, pp. 726--737, Jul 2017.

\bibitem{XiaTHZDZ17}
G.~Xia, X.~Tong, F.~Hu, Y.~Zhong, M.~Datcu, and L.~Zhang, ``Exploiting deep
  features for remote sensing image retrieval: {A} systematic investigation,''
  \emph{CoRR}, vol. abs/1707.07321, 2017.

\bibitem{deepFeature1}
W.~Zhao and S.~Du, ``Learning multiscale and deep representations for
  classifying remotely sensed imagery,'' \emph{ISPRS J. Photogrammetry and
  Remote Sensing}, vol. 113, pp. 155--165, 2016.

\bibitem{deepFeature2}
W.~Zhao, Z.~Guo, J.~Yue, X.~Zhang, and L.~Luo, ``On combining multiscale deep
  learning features for the classification of hyperspectral remote sensing
  imagery,'' \emph{International Journal of Remote Sensing}, vol.~36, no.~13,
  pp. 3368--3379, 2015.

\bibitem{deepFeature3}
C.~Zhang, X.~Pan, H.~Li, A.~Gardiner, I.~Sargent, J.~Hare, and P.~M. Atkinson,
  ``A hybrid mlp-cnn classifier for very fine resolution remotely sensed image
  classification,'' \emph{ISPRS J. Photogrammetry and Remote Sensing}, vol.
  140, pp. 133--144, 2018.

\bibitem{endToEnd1}
E.~Maggiori, Y.~Tarabalka, G.~Charpiat, and P.~Alliez, ``Convolutional neural
  networks for large-scale remote-sensing image classification,'' \emph{{IEEE}
  Trans. on Geoscience and Remote Sensing}, vol.~55, no.~2, pp. 645--657, 2017.

\bibitem{endToEnd2}
N.~Kussul, M.~Lavreniuk, S.~Skakun, and A.~Shelestov, ``Deep learning
  classification of land cover and crop types using remote sensing data,''
  \emph{{IEEE} Geoscience and Remote Sensing Letters}, vol.~14, no.~5, pp.
  778--782, 2017.

\bibitem{endToEnd3}
M.~Volpi and D.~Tuia, ``Dense semantic labeling of subdecimeter resolution
  images with convolutional neural networks,'' \emph{{IEEE} Trans. on
  Geoscience and Remote Sensing}, vol.~55, no.~2, pp. 881--893, 2017.

\bibitem{domainAdaption}
E.~Othman, Y.~Bazi, F.~Melgani, H.~Alhichri, N.~Alajlan, and M.~Zuair, ``Domain
  adaptation network for cross-scene classification,'' \emph{{IEEE} Trans. on
  Geoscience and Remote Sensing}, vol.~55, no.~8, pp. 4441--4456, 2017.

\bibitem{lu2017active}
Q.~Lu, Y.~Ma, and G.-S. Xia, ``Active learning for training sample selection in
  remote sensing image classification using spatial information,'' \emph{Remote
  Sensing Letters}, vol.~8, no.~12, pp. 1210--1219, 2017.

\bibitem{hu2015comparative}
J.~Hu, G.-S. Xia, F.~Hu, and L.~Zhang, ``A comparative study of sampling
  analysis in the scene classification of optical high-spatial resolution
  remote sensing imagery,'' \emph{Remote Sensing}, vol.~7, no.~11, pp.
  14\,988--15\,013, 2015.

\bibitem{batchSelection}
S.~Chakraborty, V.~Balasubramanian, Q.~Sun, S.~Panchanathan, and J.~Ye,
  ``Active batch selection via convex relaxations with guaranteed solution
  bounds,'' \emph{{IEEE} Trans. on Pattern Analysis and Machine Intelligence},
  vol.~37, no.~10, pp. 1945--1958, 2015.

\bibitem{dataset1}
M.~Gerke, F.~Rottensteiner, J.~D. Wegner, and G.~Sohn, ``Isprs semantic
  labeling contest,'' in \emph{Photogrammetric Computer Vision.
  \url{http://www2.isprs.org/commissions/comm3/wg4/semantic-labeling.html}},
  2014.

\bibitem{dataset2}
E.~Maggiori, Y.~Tarabalka, G.~Charpiat, and P.~Alliez, ``Can semantic labeling
  methods generalize to any city? the inria aerial image labeling benchmark,''
  in \emph{IEEE International Symposium on Geoscience and Remote Sensing},
  2017, pp. 3226--3229.

\bibitem{dataset3}
G.~Mattyus, S.~Wang, S.~Fidler, and R.~Urtasun, ``Enhancing road maps by
  parsing aerial images around the world,'' in \emph{IEEE International
  Conference on Computer Vision}, 2015, pp. 1689--1697.

\bibitem{DatasetReview}
L.~Ma, M.~Li, X.~Ma, L.~Cheng, P.~Du, and Y.~Liu, ``A review of supervised
  object-based land-cover image classification,'' \emph{ISPRS J. Photogrammetry
  and Remote Sensing}, vol. 130, pp. 277--293, 2017.

\bibitem{dataset4}
V.~Mnih, ``Machine learning for aerial image labeling,'' Ph.D. dissertation,
  University of Toronto (Canada), 2013.

\bibitem{tong2018large}
X.-Y. Tong, Q.~Lu, G.-S. Xia, and L.~Zhang, ``Large-scale land cover
  classification in gaofen-2 satellite imagery,'' \emph{arXiv preprint
  arXiv:1806.00901}, 2018.

\bibitem{spectralDrawback1}
T.~Blaschke, ``What's wrong with pixels? some recent developments interfacing
  remote sensing and gis,'' \emph{GeoBIT/GIS}, vol.~6, pp. 12--17, 2001.

\bibitem{spectralDrawback2}
C.~Burnett and T.~Blaschke, ``A multi-scale segmentation/object relationship
  modelling methodology for landscape analysis,'' \emph{Ecological Modelling},
  vol. 168, no.~3, pp. 233--249, 2003.

\bibitem{spectralDrawback3}
U.~C. Benz, P.~Hofmann, G.~Willhauck, I.~Lingenfelder, and M.~Heynen,
  ``Multi-resolution, object-oriented fuzzy analysis of remote sensing data for
  gis-ready information,'' \emph{ISPRS J. Photogrammetry and Remote Sensing},
  vol.~58, no. 3-4, pp. 239--258, 2004.

\bibitem{texture}
F.~Pacifici, M.~Chini, and W.~J. Emery, ``A neural network approach using
  multi-scale textural metrics from very high-resolution panchromatic imagery
  for urban land-use classification,'' \emph{Remote Sensing of Environment},
  vol. 113, no.~6, pp. 1276--1292, 2009.

\bibitem{XiaDG10}
G.~Xia, J.~Delon, and Y.~Gousseau, ``Shape-based invariant texture indexing,''
  \emph{International Journal of Computer Vision}, vol.~88, no.~3, pp.
  382--403, 2010.

\bibitem{XiaLBZ17}
G.~Xia, G.~Liu, X.~Bai, and L.~Zhang, ``Texture characterization using shape
  co-occurrence patterns,'' \emph{{IEEE} Trans. Image Processing}, vol.~26,
  no.~10, pp. 5005--5018, 2017.

\bibitem{shape}
L.~Zhang, X.~Huang, B.~Huang, and P.~Li, ``A pixel shape index coupled with
  spectral information for classification of high spatial resolution remotely
  sensed imagery,'' \emph{{IEEE} Trans. on Geoscience and Remote Sensing},
  vol.~44, no.~10, pp. 2950--2961, 2006.

\bibitem{structure}
D.~Tuia, F.~Ratle, A.~Pozdnoukhov, and G.~Camps-Valls, ``Multisource composite
  kernels for urban-image classification,'' \emph{{IEEE} Geoscience and Remote
  Sensing Letters}, vol.~7, no.~1, pp. 88--92, 2010.

\bibitem{Xia2010StructuralHS}
G.~Xia, W.~Yang, J.~Delon, Y.~Gousseau, H.~P.~H. Sun, and H.~Maitre,
  ``Structural high-resolution satellite image indexing,'' in \emph{ISPRS TC
  VII Symposium ¨C 100 Years ISPRS, Vienna, Austria}, 2010.

\bibitem{comparison1}
T.~Blaschke, ``Object based image analysis for remote sensing,'' \emph{ISPRS J.
  Photogrammetry and Remote Sensing}, vol.~65, no.~1, pp. 2--16, 2010.

\bibitem{comparison2}
G.~Yan, J.-F. Mas, B.~Maathuis, Z.~Xiangmin, and P.~Van~Dijk, ``Comparison of
  pixel-based and object-oriented image classification approaches¡ªa case
  study in a coal fire area, wuda, inner mongolia, china,'' \emph{International
  Journal of Remote Sensing}, vol.~27, no.~18, pp. 4039--4055, 2006.

\bibitem{comparison3}
S.~W. Myint, P.~Gober, A.~Brazel, S.~Grossman-Clarke, and Q.~Weng, ``Per-pixel
  vs. object-based classification of urban land cover extraction using high
  spatial resolution imagery,'' \emph{Remote Sensing of Environment}, vol. 115,
  no.~5, pp. 1145--1161, 2011.

\bibitem{comparison4}
D.~C. Duro, S.~E. Franklin, and M.~G. Dub{\'e}, ``A comparison of pixel-based
  and object-based image analysis with selected machine learning algorithms for
  the classification of agricultural landscapes using spot-5 hrg imagery,''
  \emph{Remote Sensing of Environment}, vol. 118, pp. 259--272, 2012.

\bibitem{deepFeature4}
S.~Paisitkriangkrai, J.~Sherrah, P.~Janney, V.-D. Hengel \emph{et~al.},
  ``Effective semantic pixel labelling with convolutional networks and
  conditional random fields,'' in \emph{IEEE Conference on Computer Vision and
  Pattern Recognition Workshops}, 2015, pp. 36--43.

\bibitem{deepFeature5}
N.~Audebert, B.~Le~Saux, and S.~Lefevre, ``How useful is region-based
  classification of remote sensing images in a deep learning framework?'' in
  \emph{{IEEE} International Geoscience and Remote Sensing Symposium}, 2016,
  pp. 5091--5094.

\bibitem{deepFeature6}
S.~Paisitkriangkrai, J.~Sherrah, P.~Janney, and A.~van~den Hengel, ``Semantic
  labeling of aerial and satellite imagery,'' \emph{{IEEE} J. Selected Topics
  in Applied Earth Observations and Remote Sensing}, vol.~9, no.~7, pp.
  2868--2881, 2016.

\bibitem{endToEnd4}
E.~Maggiori, Y.~Tarabalka, G.~Charpiat, and P.~Alliez, ``High-resolution
  semantic labeling with convolutional neural networks,'' \emph{arXiv preprint
  arXiv:1611.01962}, 2016.

\bibitem{endToEnd5}
Y.~Liu, D.~Minh~Nguyen, N.~Deligiannis, W.~Ding, and A.~Munteanu,
  ``Hourglass-shapenetwork based semantic segmentation for high resolution
  aerial imagery,'' \emph{Remote Sensing}, vol.~9, no.~6, p. 522, 2017.

\bibitem{WODS1}
J.~Sherrah, ``Fully convolutional networks for dense semantic labelling of
  high-resolution aerial imagery,'' \emph{arXiv preprint arXiv:1606.02585},
  2016.

\bibitem{WODS2}
C.~Persello and A.~Stein, ``Deep fully convolutional networks for the detection
  of informal settlements in vhr images,'' \emph{{IEEE} Geoscience and Remote
  Sensing Letters}, vol.~14, no.~12, pp. 2325--2329, 2017.

\bibitem{Segnet}
V.~Badrinarayanan, A.~Kendall, and R.~Cipolla, ``Segnet: A deep convolutional
  encoder-decoder architecture for image segmentation,'' \emph{{IEEE} Trans. on
  Pattern Analysis and Machine Intelligence}, vol.~39, no.~12, pp. 2481--2495,
  2017.

\bibitem{DeepLab}
L.~C. Chen, G.~Papandreou, I.~Kokkinos, K.~Murphy, and A.~L. Yuille, ``Deeplab:
  Semantic image segmentation with deep convolutional nets, atrous convolution,
  and fully connected crfs,'' \emph{{IEEE} Trans. on Pattern Analysis and
  Machine Intelligence}, vol.~40, no.~4, pp. 834--848, 2018.

\bibitem{OCNN}
C.~Zhang, I.~Sargent, X.~Pan, H.~Li, A.~Gardiner, J.~Hare, and P.~M. Atkinson,
  ``An object-based convolutional neural network (ocnn) for urban land use
  classification,'' \emph{Remote Sensing of Environment}, vol. 216, pp. 57--70,
  2018.

\bibitem{TLdefine}
X.~Li, L.~Zhang, B.~Du, L.~Zhang, and Q.~Shi, ``Iterative reweighting
  heterogeneous transfer learning framework for supervised remote sensing image
  classification,'' \emph{{IEEE} J. Selected Topics in Applied Earth
  Observations and Remote Sensing}, vol.~10, no.~5, pp. 2022--2035, 2017.

\bibitem{InvariantFeature1}
E.~Izquierdo-Verdiguier, V.~Laparra, L.~Gomez-Chova, and G.~Camps-Valls,
  ``Encoding invariances in remote sensing image classification with svm,''
  \emph{{IEEE} Geoscience and Remote Sensing Letters}, vol.~10, no.~5, pp.
  981--985, 2013.

\bibitem{InvariantFeature2}
L.~Bruzzone and C.~Persello, ``A novel approach to the selection of spatially
  invariant features for the classification of hyperspectral images with
  improved generalization capability,'' \emph{{IEEE} Trans. on Geoscience and
  Remote Sensing}, vol.~47, no.~9, pp. 3180--3191, 2009.

\bibitem{DistributionAdaptation1}
D.~Tuia and G.~Camps-Valls, ``Kernel manifold alignment for domain
  adaptation,'' \emph{Public Library of Science}, vol.~11, no.~2, p. e0148655,
  2016.

\bibitem{DistributionAdaptation2}
H.~L. Yang and M.~M. Crawford, ``Domain adaptation with preservation of
  manifold geometry for hyperspectral image classification,'' \emph{{IEEE} J.
  Selected Topics in Applied Earth Observations and Remote Sensing}, vol.~9,
  no.~2, pp. 543--555, 2016.

\bibitem{DistributionAdaptation3}
G.~Jun and J.~Ghosh, ``Spatially adaptive classification of land cover with
  remote sensing data,'' \emph{IEEE Transactions on Geoscience and Remote
  Sensing}, vol.~49, no.~7, pp. 2662--2673, 2011.

\bibitem{ActiveLearning1}
C.~Persello and L.~Bruzzone, ``Active learning for domain adaptation in the
  supervised classification of remote sensing images,'' \emph{{IEEE} Trans. on
  Geoscience and Remote Sensing}, vol.~50, no.~11, pp. 4468--4483, 2012.

\bibitem{ActiveLearning2}
B.~Demir, F.~Bovolo, and L.~Bruzzone, ``Detection of land-cover transitions in
  multitemporal remote sensing images with active-learning-based compound
  classification,'' \emph{{IEEE} Trans. on Geoscience and Remote Sensing},
  vol.~50, no.~5, pp. 1930--1941, 2012.

\bibitem{ActiveLearning3}
B.~Demir, L.~Minello, and L.~Bruzzone, ``Definition of effective training sets
  for supervised classification of remote sensing images by a novel
  cost-sensitive active learning method,'' \emph{{IEEE} Trans. on Geoscience
  and Remote Sensing}, vol.~52, no.~2, pp. 1272--1284, 2014.

\bibitem{SSLdefine}
C.~Persello and L.~Bruzzone, ``Active and semisupervised learning for the
  classification of remote sensing images,'' \emph{{IEEE} Trans. on Geoscience
  and Remote Sensing}, vol.~52, no.~11, pp. 6937--6956, 2014.

\bibitem{ClassifierAdaptation1}
L.~Bruzzone, M.~Chi, and M.~Marconcini, ``A novel transductive svm for
  semisupervised classification of remote-sensing images,'' \emph{{IEEE} Trans.
  on Geoscience and Remote Sensing}, vol.~44, no.~11, pp. 3363--3373, 2006.

\bibitem{ClassifierAdaptation2}
L.~G{\'o}mez-Chova, G.~Camps-Valls, J.~Munoz-Mari, and J.~Calpe,
  ``Semisupervised image classification with laplacian support vector
  machines,'' \emph{{IEEE} Geoscience and Remote Sensing Letters}, vol.~5,
  no.~3, pp. 336--340, 2008.

\bibitem{ClassifierAdaptation3}
G.~Matasci, M.~Volpi, M.~Kanevski, L.~Bruzzone, and D.~Tuia, ``Semisupervised
  transfer component analysis for domain adaptation in remote sensing image
  classification,'' \emph{{IEEE} Trans. on Geoscience and Remote Sensing},
  vol.~53, no.~7, pp. 3550--3564, 2015.

\bibitem{DL}
Y.~LeCun, Y.~Bengio, and G.~Hinton, ``Deep learning,'' \emph{Nature}, vol. 521,
  no. 7553, p. 436, 2015.

\bibitem{DeepFE1}
D.~Marmanis, M.~Datcu, T.~Esch, and U.~Stilla, ``Deep learning earth
  observation classification using imagenet pretrained networks,'' \emph{{IEEE}
  Geoscience and Remote Sensing Letters}, vol.~13, no.~1, pp. 105--109, 2016.

\bibitem{DeepFE2}
B.~Zhao, B.~Huang, and Y.~Zhong, ``Transfer learning with fully pretrained deep
  convolution networks for land-use classification,'' \emph{{IEEE} Geoscience
  and Remote Sensing Letters}, vol.~14, no.~9, pp. 1436--1440, 2017.

\bibitem{JointFinetune2}
W.~Ge and Y.~Yu, ``Borrowing treasures from the wealthy: Deep transfer learning
  through selective joint fine-tuning,'' in \emph{IEEE Conference on Computer
  Vision and Pattern Recognition}, vol.~6, 2017.

\bibitem{STDCNN}
B.~Huang, B.~Zhao, and Y.~Song, ``Urban land-use mapping using a deep
  convolutional neural network with high spatial resolution multispectral
  remote sensing imagery,'' \emph{Remote Sensing of Environment}, vol. 214, pp.
  73--86, 2018.

\bibitem{ResNet}
K.~He, X.~Zhang, S.~Ren, and J.~Sun, ``Deep residual learning for image
  recognition,'' in \emph{IEEE Conference on Computer Vision and Pattern
  Recognition}, 2016, pp. 770--778.

\bibitem{PseudoLabel1}
K.~Wu and K.-H. Yap, ``Fuzzy svm for content-based image retrieval: a
  pseudo-label support vector machine framework,'' \emph{{IEEE} Computational
  Intelligence Magazine}, vol.~1, no.~2, pp. 10--16, 2006.

\bibitem{PseudoLabel2}
D.-H. Lee, ``Pseudo-label: The simple and efficient semi-supervised learning
  method for deep neural networks,'' in \emph{Workshop on Challenges in
  Representation Learning, ICML}, vol.~3, 2013, p.~2.

\bibitem{JointFinetune1}
Y.~Xue, X.~Liao, L.~Carin, and B.~Krishnapuram, ``Multi-task learning for
  classification with dirichlet process priors,'' \emph{Journal of Machine
  Learning Research}, vol.~8, no. Jan, pp. 35--63, 2007.

\bibitem{weight}
Q.~Lu, X.~Huang, J.~Li, and L.~Zhang, ``A novel mrf-based multifeature fusion
  for classification of remote sensing images,'' \emph{{IEEE} Geoscience and
  Remote Sensing Letters}, vol.~13, no.~4, pp. 515--519, 2016.

\bibitem{selectiveSearch}
J.~R. Uijlings, K.~E. Van De~Sande, T.~Gevers, and A.~W. Smeulders, ``Selective
  search for object recognition,'' \emph{International Journal of Computer
  Vision}, vol. 104, no.~2, pp. 154--171, 2013.

\bibitem{graphSegmentation}
P.~F. Felzenszwalb and D.~P. Huttenlocher, ``Efficient graph-based image
  segmentation,'' \emph{International Journal of Computer Vision}, vol.~59,
  no.~2, pp. 167--181, 2004.

\bibitem{ImageNET}
J.~Deng, W.~Dong, R.~Socher, L.-J. Li, K.~Li, and L.~Fei-Fei, ``Imagenet: A
  large-scale hierarchical image database,'' in \emph{IEEE Conference on
  Computer Vision and Pattern Recognition}, 2009, pp. 248--255.

\bibitem{GLCM}
R.~M. Haralick, K.~Shanmugam \emph{et~al.}, ``Textural features for image
  classification,'' \emph{{IEEE} Trans. on Systems, Man, and Cybernetics},
  no.~6, pp. 610--621, 1973.

\bibitem{DMP}
J.~A. Benediktsson, J.~A. Palmason, and J.~R. Sveinsson, ``Classification of
  hyperspectral data from urban areas based on extended morphological
  profiles,'' \emph{{IEEE} Trans. on Geoscience and Remote Sensing}, vol.~43,
  no.~3, pp. 480--491, 2005.

\bibitem{LBP}
T.~Ojala, M.~Pietik\"ainen, and T.~Maenpaa, ``Multiresolution gray-scale and
  rotation invariant texture classification with local binary patterns,''
  \emph{{IEEE} Trans. on Pattern Analysis and Machine Intelligence}, vol.~24,
  no.~7, pp. 971--987, 2002.

\bibitem{accuracy}
P.~Olofsson, G.~M. Foody, M.~Herold, S.~V. Stehman, C.~E. Woodcock, and M.~A.
  Wulder, ``Good practices for estimating area and assessing accuracy of land
  change,'' \emph{Remote Sensing of Environment}, vol. 148, pp. 42--57, 2014.

\bibitem{limitation}
Q.~Hu, W.~Wu, T.~Xia, Q.~Yu, P.~Yang, Z.~Li, and Q.~Song, ``Exploring the use
  of google earth imagery and object-based methods in land use/cover mapping,''
  \emph{Remote Sensing}, vol.~5, no.~11, pp. 6026--6042, 2013.

\bibitem{multi-temporal1}
Y.~Shao, R.~S. Lunetta, B.~Wheeler, J.~S. Iiames, and J.~B. Campbell, ``An
  evaluation of time-series smoothing algorithms for land-cover classifications
  using modis-ndvi multi-temporal data,'' \emph{Remote Sensing of Environment},
  vol. 174, pp. 258--265, 2016.

\bibitem{multi-temporal2}
Y.~Sheng, C.~Song, J.~Wang, E.~A. Lyons, B.~R. Knox, J.~S. Cox, and F.~Gao,
  ``Representative lake water extent mapping at continental scales using
  multi-temporal landsat-8 imagery,'' \emph{Remote Sensing of Environment},
  vol. 185, pp. 129--141, 2016.

\bibitem{multi-temporal3}
N.~Kussul, S.~Skakun, A.~Shelestov, M.~Lavreniuk, B.~Yailymov, and O.~Kussul,
  ``Regional scale crop mapping using multi-temporal satellite imagery,''
  \emph{The International Archives of Photogrammetry, Remote Sensing and
  Spatial Information Sciences}, vol.~40, no.~7, p.~45, 2015.

\bibitem{multi-temporal4}
F.~Vuolo, M.~Neuwirth, M.~Immitzer, C.~Atzberger, and W.-T. Ng, ``How much does
  multi-temporal sentinel-2 data improve crop type classification?''
  \emph{International journal of applied earth observation and geoinformation},
  vol.~72, pp. 122--130, 2018.

\end{thebibliography}

\end{document}